%% file: main.tex
\documentclass{article}
\usepackage{graphicx}
\usepackage{amsmath}
\usepackage{booktabs}
\usepackage{hyperref}
\usepackage{subcaption} %
\usepackage{url}
\usepackage{xcolor}
\usepackage{booktabs}
\usepackage{adjustbox}
\usepackage{longtable}
\usepackage{listings}

\usepackage[a4paper, margin=1in]{geometry}  %

\definecolor{dkgreen}{rgb}{0,0.6,0}
\definecolor{gray}{rgb}{0.5,0.5,0.5}
\definecolor{mauve}{rgb}{0.58,0,0.82}

\usepackage{xcolor}
\usepackage{mleftright}
\usepackage{amsthm}
\usepackage[shortlabels]{enumitem}
\usepackage[capitalize,noabbrev]{cleveref}
\usepackage{pifont}
\usepackage{multirow}
\usepackage{float}
\usepackage{wrapfig}
\usepackage{xspace}

\usepackage{mathtools}

\newcommand{\codeforces}{\textsc{CodeForces}\xspace}

\title{Competitive Programming with Large Reasoning Models}
\author{OpenAI\thanks{Contributions listed in Appendix~\ref{app:authors}}
}
\date{}

\begin{document}
\maketitle

\begin{abstract}
We show that reinforcement learning applied to large language models (LLMs) significantly boosts performance on complex coding and reasoning tasks. Additionally, we compare two general-purpose reasoning models — OpenAI o1 and an early checkpoint of o3 — with a domain-specific system, o1-ioi, which uses hand-engineered inference strategies designed for competing in the 2024 International Olympiad in Informatics (IOI). We competed live at IOI 2024 with o1-ioi and, using hand-crafted test-time strategies, placed in the 49th percentile. Under relaxed competition constraints, o1-ioi achieved a gold medal. However, when evaluating later models such as o3, we find that o3 achieves gold without hand-crafted domain-specific strategies or relaxed constraints. Our findings show that although specialized pipelines such as o1-ioi yield solid improvements, the scaled-up, general-purpose o3 model surpasses those results without relying on hand-crafted inference heuristics. Notably, o3 achieves a gold medal at the 2024 IOI and obtains a \codeforces rating on par with elite human competitors. Overall, these results indicate that scaling general-purpose reinforcement learning, rather than relying on domain-specific techniques, offers a robust path toward state-of-the-art AI in reasoning domains, such as competitive programming.

\end{abstract}

\section{Introduction}

Competitive programming is widely recognized as a challenging benchmark for evaluating reasoning and coding proficiency~\cite{chen2021evaluating}.
Solving complex algorithmic problems demands advanced computational thinking and problem solving skills. Moreover, these problems are also objectively gradable, making it an ideal testbed to assess the reasoning capabilities of AI systems.

Recent work on program synthesis with large language models~\cite{austin2021program} has demonstrated that even relatively general models, ranging from 244M to 137B parameters, can generate short Python scripts from natural language instructions. Importantly, performance improves log-linearly with model size, and fine-tuning significantly boosts accuracy. Concurrently, Codex~\cite{chen2021evaluating}, an early code-focused LLM, excelled at Python program generation and powered GitHub Copilot. Further progress came from AlphaCode~\cite{li2022competition}, which tackled competitive programming tasks using large-scale code generation and heuristics at inference, and the subsequent AlphaCode2\cite{leblond2023alphacode}, whose improvements nearly doubled AlphaCode’s solved problems and placed it in the 85th percentile on the \codeforces platform.
Both AlphaCode systems used large-scale sampling of up to a million candidate solutions per problem before selecting their top 10 submissions with a hand-engineered test-time strategy.

Since then, significant progress has been made in harnessing reinforcement learning to improve LLMs’ reasoning skills. This has led to the emergence of large reasoning models (LRMs): language models trained via reinforcement learning to ``reason" and ``think through" extended chains of thought.  In particular, OpenAI’s o1~\cite{jaech2024openai,openai_learning_reason_llms} and its soon-to-be-released successor o3~\cite{openaio3} use chain-of-thought reasoning to tackle intricate tasks such as mathematics and coding. Work by DeepSeek-R1~\cite{deepseekai2025deepseekr1incentivizingreasoningcapability} and Kimi k1.5~\cite{kimiteam2025kimik15scalingreinforcement} independently illustrates how learning chain-of-thought boosts performance on both mathematical and programming challenges.

An open question is how domain-specific, hand-engineered inference strategies compare to learned approaches that models generate and execute on their own. We have three systems available that can shed light on this question: o1, o1-ioi, and early checkpoints of o3. OpenAI o1 was the first large reasoning model and used general purpose methods to improve programming performance.  Building on this foundation, o1-ioi was a fine-tuned system tailored to compete in the 2024 International Olympiad in Informatics (IOI) and used test-time strategies similar to those used in the AlphaCode system. This specialization led to strong performance improvements on both the 2024 IOI and competitive programming platforms such as \codeforces. Subsequent advances led to the development of o3, which has significantly advanced the reasoning capabilities of AI models. Unlike o1-ioi or AlphaCode, o3 does not depend on coding-specific test-time strategies defined by humans. Instead, we found that complex test-time reasoning strategies emerged naturally from end-to-end RL, leading to unprecedented performance on competitive programming benchmarks.

This report provides a high-level overview of the importance of reasoning in coding tasks such as competitive programming, the progress of OpenAI’s large reasoning models in programming ability, and our evaluation methodology and results on various competitive programming and coding benchmarks.

\section{OpenAI o1}
We start with OpenAI o1, a large language model trained with reinforcement learning to tackle complex reasoning tasks. By generating an extended internal chain of thought before answering~\cite{wei2022chain}, o1 resembles a human who methodically works through a challenging problem step by step. Reinforcement learning refines this chain-of-thought process, helping the model identify and correct errors, break down complex tasks into manageable parts, and explore alternate solution paths when an approach fails. These in-context reasoning capabilities substantially boost o1’s overall performance on a wide range of tasks.

Additionally, OpenAI o1 is trained to use external tools~\cite{schick2023toolformer}, especially for writing and executing code in a secure environment.\footnote{\url{https://platform.openai.com/docs/assistants/tools/code-interpreter}} This capability lets o1 verify whether its generated code compiles, passes provided test cases, and meets other correctness checks. By testing and refining its outputs, o1 iteratively improves its solutions over the course of a single sample.

\subsection{\codeforces Benchmark}
\codeforces is a programming competition website that hosts live contests.
It is internationally competitive and frequented by some of the best competitive programmers in the world.

To assess our models' competitive programming abilities, we simulated \codeforces contests under conditions that closely mirrored real competitions. This included using the full test suite for each problem and enforcing appropriate time and memory constraints for solutions.

Our evaluation focused on Division 1 contests from 2024 and December 2023, ensuring all test contests occurred after the data cut-off for both pretraining and RL. Additionally, we conducted a contamination check as a sanity measure, leveraging the OpenAI embedding API to verify that test problems had not been seen during training.

\begin{figure}[h!]
    \centering
    \includegraphics[width=0.85\textwidth]{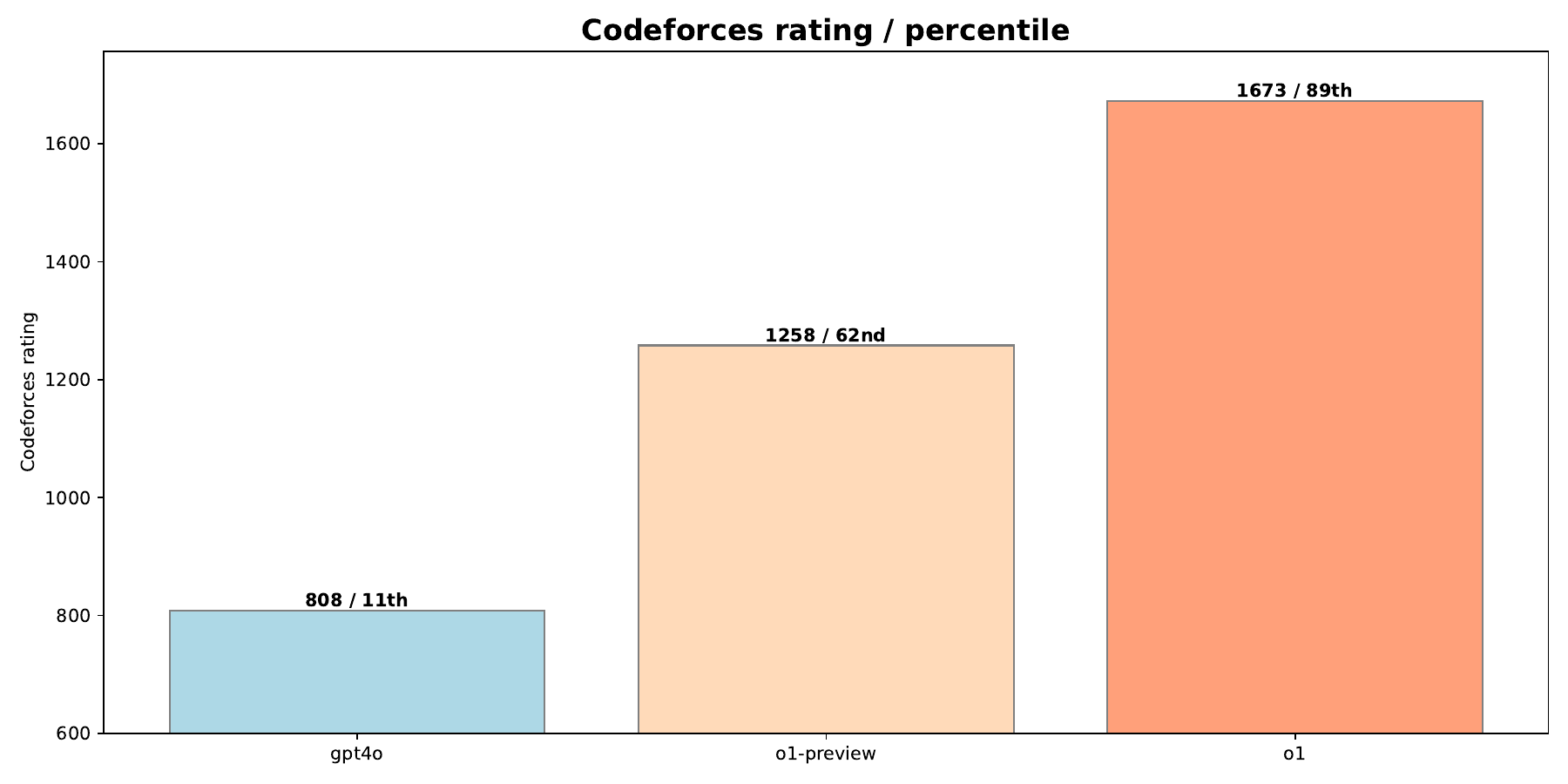}
    \caption{Comparing reasoning LLMs OpenAI o1-preview and o1 to gpt-4o on \codeforces.}
    \label{fig:codeforces_gpt4o}
\end{figure}

We compared o1 against a non-reasoning LLM (gpt-4o) and an earlier reasoning model (o1-preview). Figure~\ref{fig:codeforces_gpt4o} shows how both o1-preview and o1 dramatically outperform gpt-4o, highlighting the effectiveness of reinforcement learning for complex reasoning. The o1-preview model achieved a \codeforces rating of 1258 (62nd percentile) --- up from gpt-4o’s 808 (11th percentile). Further training pushed o1’s rating to 1673 (89th percentile), establishing a new milestone for AI performance in competitive programming.

In Appendix \ref{app:cf_details} we provide additional details of which problems our models can solve and how ratings were calculated.  

\section{OpenAI o1-ioi}
During our development and evaluation of OpenAI o1, we found that increasing both the amount of reinforcement learning (RL) compute and test-time inference compute consistently improved model performance.

\begin{figure}[h!] \centering \includegraphics[width=0.85\textwidth]{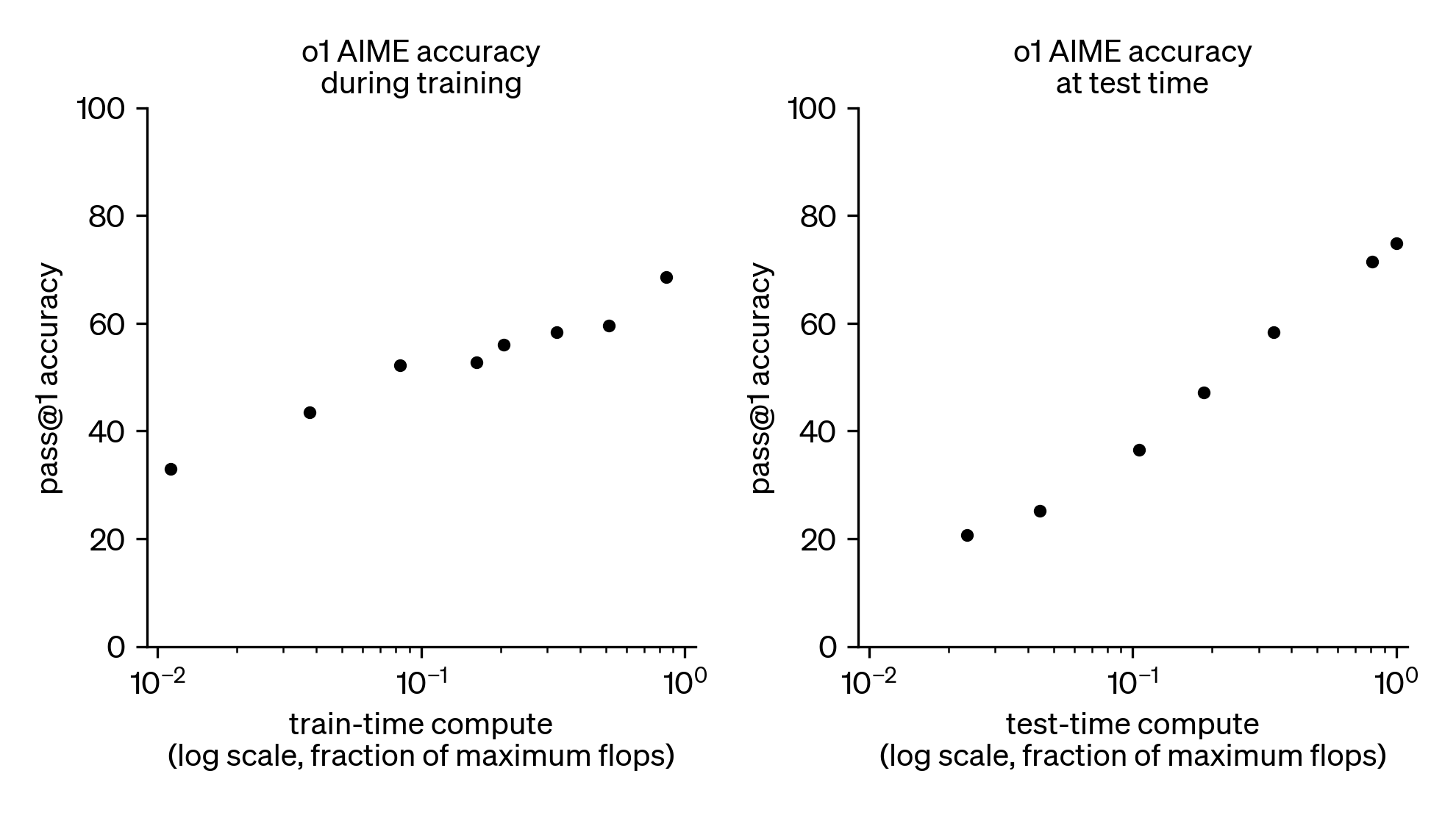} \caption{Additional RL training and additional test-time compute improves competitive mathematics performance.} \label{fig:aime_train_test} \end{figure}

As shown in Figure~\ref{fig:aime_train_test}, scaling RL training and extending test-time inference led to marked gains, highlighting the importance of optimizing these two compute dimensions to push performance beyond conventional LLM pretraining.

Building on these insights, we created the \textit{o1-ioi} system for competing at the 2024 International Olympiad in Informatics (IOI). In addition to continued RL training targeted at coding tasks, \textit{o1-ioi} incorporates specialized test-time inference strategies engineered for competitive programming.

\subsection{Coding RL Fine-tuning}
Our first step extended the reinforcement learning phase of OpenAI o1, focusing on coding tasks. By dedicating additional training compute to programming problems, we bolstered the model’s ability to plan, implement, and debug more involved solutions. Concretely:

\begin{enumerate}
    \item We resumed RL training from the OpenAI o1 checkpoint.
    \item We specifically emphasized challenging programming problems, helping the model improve C++ generation and runtime checks.
    \item We guided the model to produce outputs in the IOI submission format.
\end{enumerate}

This added focus on coding allowed o1-ioi to write and execute C++ programs during inference. The model improved its reasoning by iteratively running and refining solutions, thereby strengthening both its coding and problem-solving skills.

\subsection{o1-ioi Test-time Strategy}
\label{section:test-time-strategy}
At a high level, we divided each IOI problem into its constituent subtasks, sampled 10,000 solutions from o1-ioi for each subtask, and then employed a clustering- and reranking-based approach to decide which solutions from this set to submit.

\paragraph{Problem formulation} For o1-ioi we chose to attempt to solve the individual subtasks of each problem separately, as the scoring for IOI is done on a subtask-by-subtask basis and gives each competitor the maximum score over all of their attempts on each subtask. To do this, we divided each IOI problem into its composite subtasks (using the divisions laid out in the scoring guide for each problem). This was done simply by creating one version of the document for each subtask with the information about the other subtasks removed.

\paragraph{Clustering}
We clustered the generated solutions based on their outputs on model-generated test inputs. For each subtask, we first prompted the model to write random test input generators in C++ given the problem specification and subtask. We used these generators to generate 256 random test inputs. To ensure the validity of these test inputs, we then prompted the model to write test input validators in C++ that check, given a test input, whether it satisfies the subtask constraints. Finally, we accepted each test input that passes at least 75\% of the validators. For each subtask, we generated 256 of these random test case inputs, and then clustered based on their outputs for these test cases. Any programs that matched each other's outputs on all test inputs would be placed in the same cluster.

\paragraph{Reranking}
We then implemented the reranking core of our test-time compute strategy. We scored each solution based on:
\begin{itemize}
    \item The quality of the solution according to a learned scoring function.
    \item Errors on model-generated test inputs.
    \item Failing the provided public test cases.
\end{itemize}
Each cluster was given a score defined as the average score of the samples it contained minus a penalty for each time a sample submission was attempted from that cluster. The weights of all of these penalties were tuned by random search on solutions to previous years' IOI problems, by directly simulating the submission process.

\paragraph{Submission}
We then submitted up to \textbf{50} (the maximum number allowed for human competitors) of these solutions in a round-robin fashion over subtasks, starting from the hardest. We selected the top-ranked solution in the top-ranked cluster for each given subtask. When a subtask was solved (meaning that the maximum score was attained), we ceased sampling on that subtask. When submitting solutions to any subtask that was a strict superset of a solved subtask, we would filter out any solutions that did not match the outputs on test inputs of the solved constituent subtasks, allowing us to rapidly narrow down candidate solutions on harder subtasks by rejecting those that would almost certainly have failed easier subtasks.

\subsection{\codeforces Benchmark}
Once again, we simulated \codeforces contests to evaluate o1-ioi's coding abilities, closely mirroring contest conditions with the complete test suite for each problem and appropriate time and memory restrictions for solutions.

\begin{figure}[h!] \centering \includegraphics[width=0.85\textwidth]{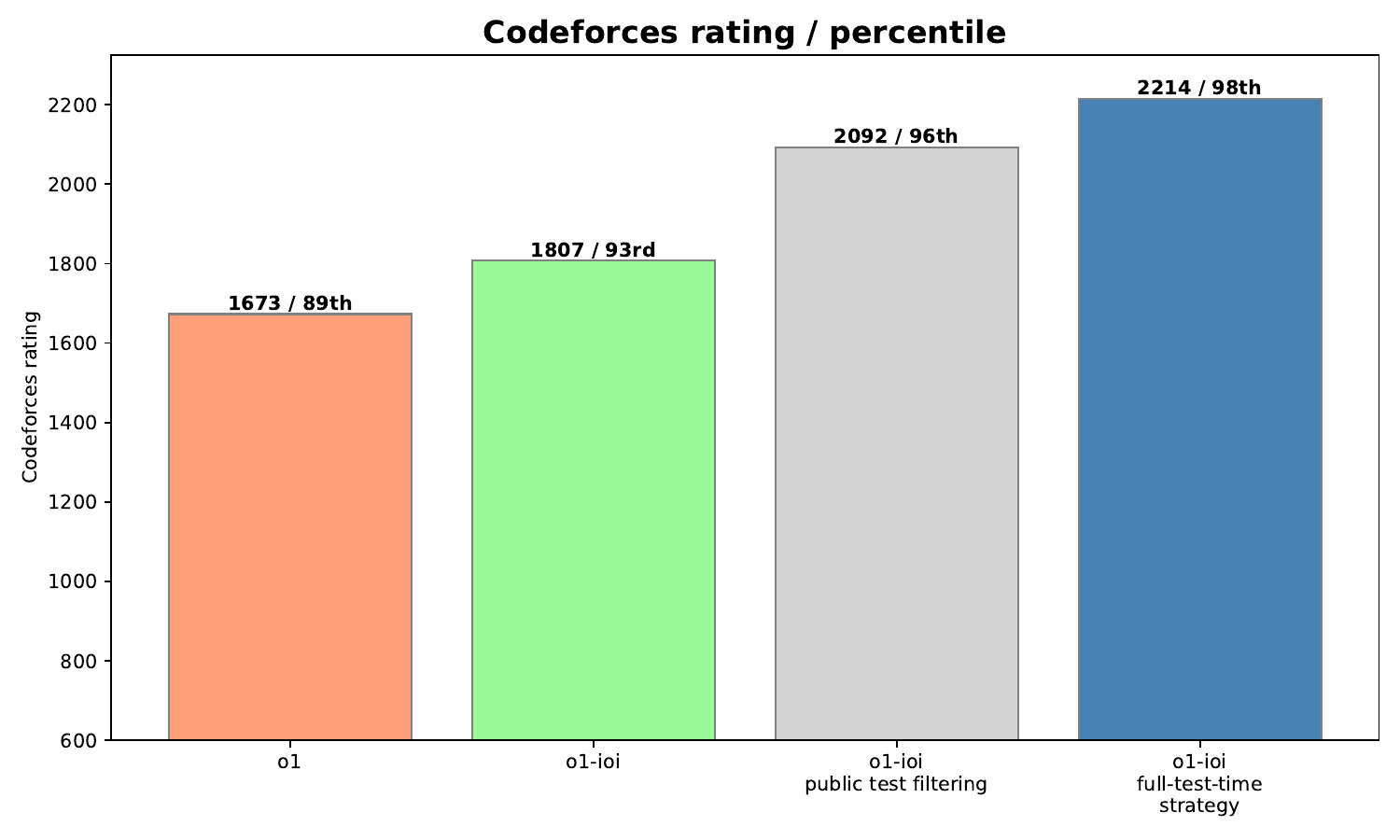} \caption{Further training OpenAI o1 on coding tasks and incorporating test-time strategies improves performance.} \label{fig:codeforces_ioi} \end{figure}

Figure~\ref{fig:codeforces_ioi} shows that o1-ioi reached a \codeforces rating of 1807, outperforming 93\% of competitors --- demonstrating clear improvements from additional RL training on coding tasks. When we applied a simple filter rejecting any solution that failed public tests, the rating rose to 2092 (96th percentile). Our complete test-time strategy pushed performance even further, attaining a rating of 2214 (98th percentile). These results confirm that domain-specific RL fine-tuning paired with advanced selection heuristics can significantly boost competitive programming outcomes.

\subsection{IOI 2024 Live Competition}

\begin{figure}[h!] \centering \includegraphics[width=0.85\textwidth]{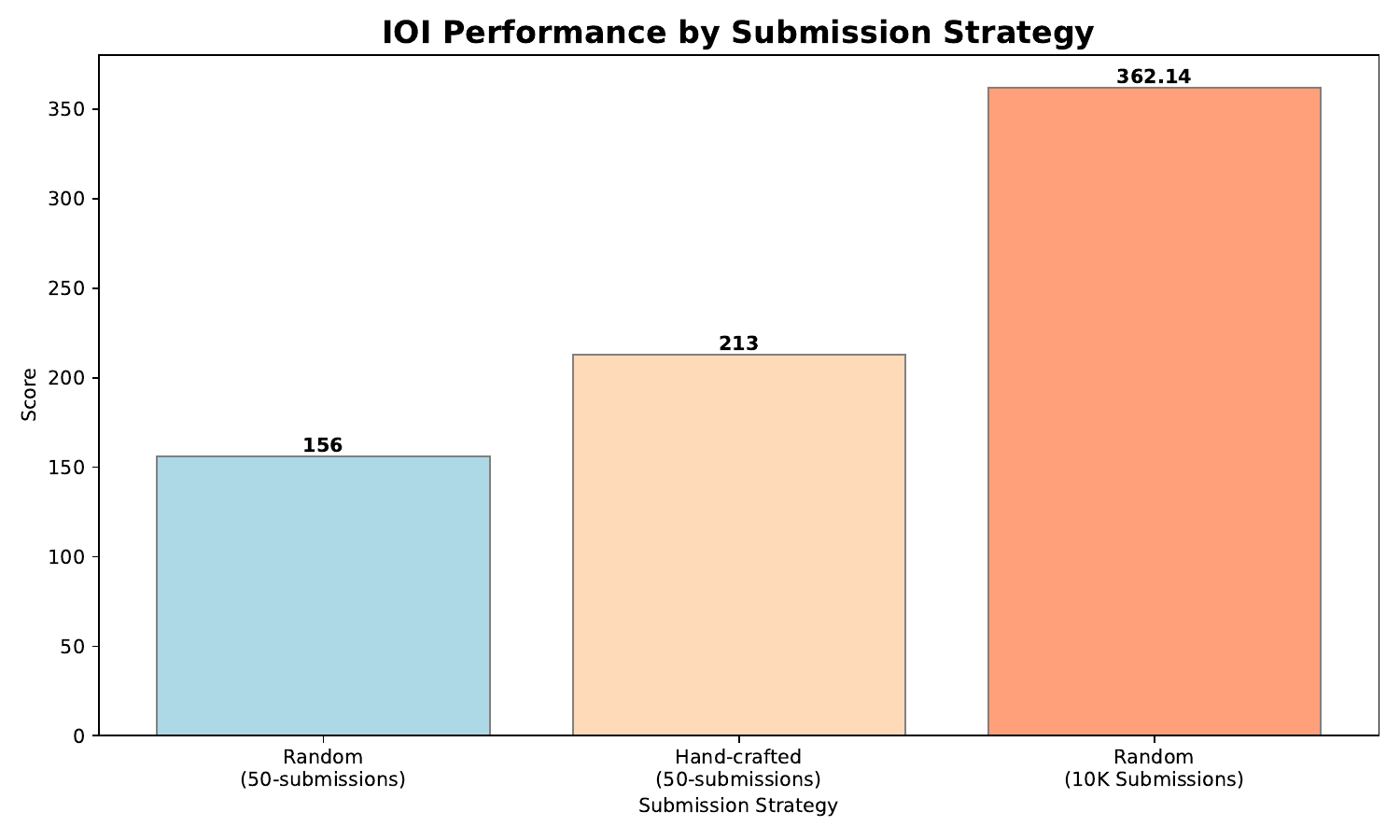} \caption{Performance of o1-ioi competing at IOI 2024.} \label{fig:ioi_2024} \end{figure}

The o1-ioi system participated in the 2024 International Olympiad in Informatics (IOI) under the same conditions as human contestants. It had ten hours to solve six challenging algorithmic problems and was allowed up to 50 submissions per problem. We show the results in Figure~\ref{fig:ioi_2024}.

During the competition, our system generated 10,000 candidate solutions for each problem, and selected 50 submissions using our test-time selection strategy. This strategy prioritized submissions based on their performance on the IOI public test cases, model-generated test cases, and a learned scoring function. The model scored 213 points, placing it in the 49th percentile of the competition.

In comparison, selecting 50 random submissions would have yielded an average score of only 156 points, indicating that the selection strategy contributed nearly 60 additional points under the competition's constraints.

When the submission limit was relaxed to 10,000 per problem, the model's performance improved dramatically. Without employing any test-time selection strategy, it achieved a score of 362.14, surpassing the gold medal threshold and demonstrating the model's potential. We show samples that yielded the 362.14 score in Appendix~\ref{appendix:ioi_samples}.

\section{OpenAI o3}
Building on the insights gained from o1 and o1-ioi, we explore the limits of reinforcement learning (RL) training alone, without relying on human-engineered test-time strategies. While o1-ioi achieved strong results by combining additional RL fine-tuning with carefully designed test-time inference pipelines, its success hinged on human intervention to define and implement these strategies. We sought to explore the performance of a model even further trained with RL with the ability to autonomously develop and execute its own test-time reasoning strategies. To this end, we obtained access to early checkpoints of o3~\cite{openaio3} to evaluate on competitive programming tasks. 

\subsection{\codeforces Benchmark}
We evaluate an early checkpoint of the o3 model on our \codeforces benchmark set, where each prompt includes the problem description, constraints, and any available sample test cases.

\begin{figure}[h!]
\centering
\includegraphics[width=0.85\textwidth]{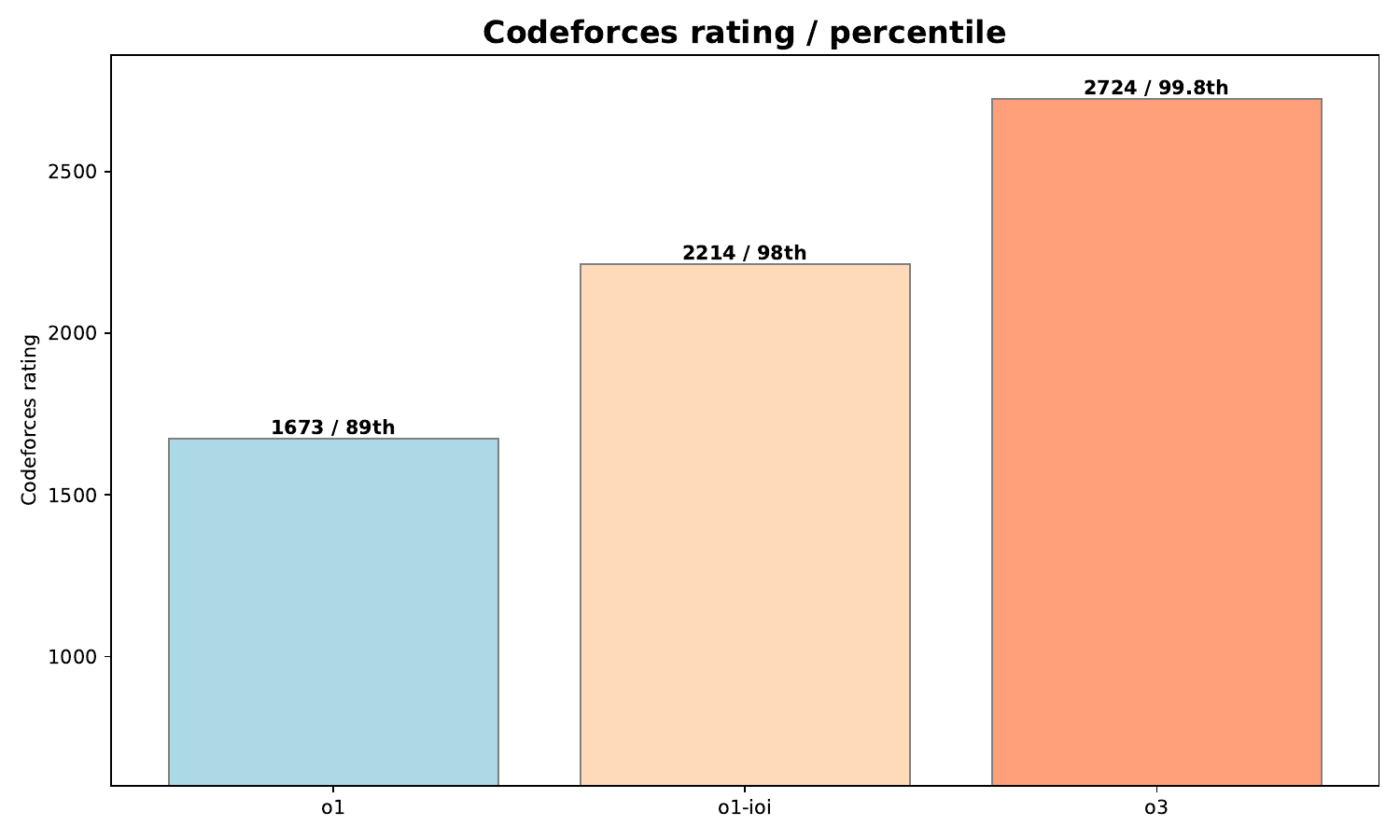}
\caption{Performance of OpenAI o3 on the \codeforces benchmark.}
\label{fig:o3_codeforces}
\end{figure}

As shown in Figure~\ref{fig:o3_codeforces}, further RL training provided a significant improvement over both o1 and the full o1-ioi system. Notably, the transition from the o1-ioi model to o3 resulted in a rating increase from 2214 (98th percentile) to 2724 (99.8th percentile), reflecting a substantial leap in competitive programming performance. This improvement demonstrates o3’s ability to solve a wider range of complex algorithmic problems with higher reliability, pushing its capabilities closer to top-tier human competitors on \codeforces.

\begin{figure}[h!]
\centering
\includegraphics[width=0.95\textwidth]{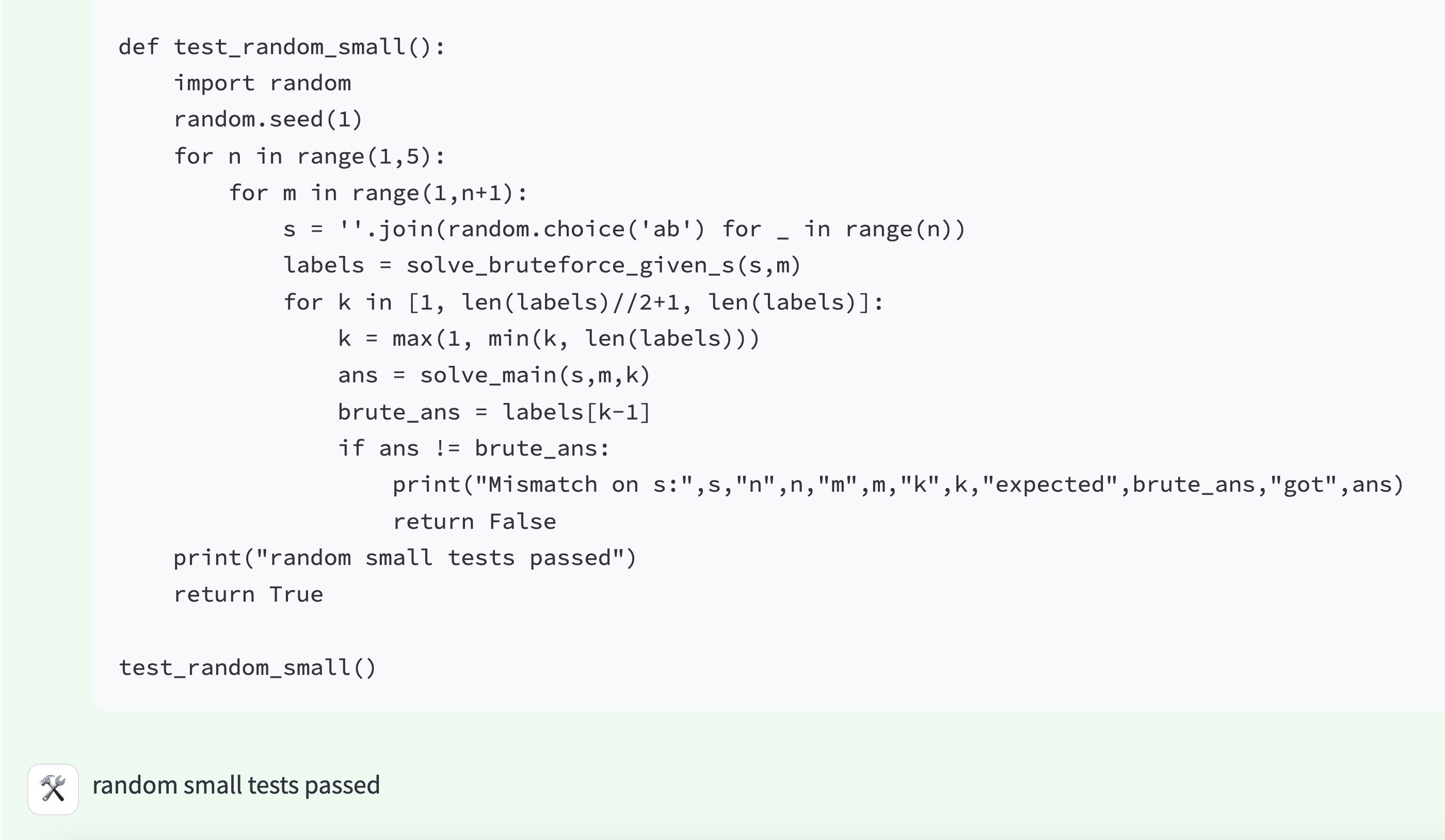}
\caption{o3 testing its own solution. This reflects a sophisticated reasoning strategy that partially implements the hand-designed test-time strategy used for o1-ioi in IOI 2024.}
\label{fig:o3_strategy}
\end{figure}

In addition to its significantly improved problem-solving capabilities, we observe that o3 demonstrates more insightful and deliberate chains of thought. The model not only writes and executes code to validate its solutions against public test cases, it also refines its approach based on these verifications. Figure~\ref{fig:o3_strategy} shows an advanced test-time strategy discovered by o3: for problems where verification is nontrivial, it often writes simple brute-force solutions --- trading efficiency for correctness --- then cross-checks the outputs against its more optimized algorithmic implementations. This self-imposed validation mechanism lets o3 catch potential errors and improve the reliability of its solutions.

\subsection{IOI 2024 Benchmark}
Although we competed in IOI 2024 using o1-ioi, we retrospectively evaluated a checkpoint of o3 on the same six IOI 2024 problems to compare performance under identical conditions. As with o1-ioi, we strictly adhered to the official IOI rules, which permit a maximum of 50 submissions per problem.

The o3 results on the IOI 2024 were produced by a later version of o3 than the \codeforces results, and included additional fresher training data. 
IOI 2024 occurred after the training cut-off for this model, and we additionally confirmed with search that the IOI test problems are not contaminated with the training set.

\paragraph{Sampling Approach.}
Unlike o1-ioi, which sampled solutions separately for each subtask, we adopted a different approach when evaluating o3: \textit{sampling from a single prompt containing the original problem statement}. Additionally, while o1-ioi generated 10K solutions per subtask, for o3 we sampled only 1K solutions per problem.

Selection strategies also differed between the two models. Whereas o1-ioi relied on a complex, human-defined test-time strategy (\ref{section:test-time-strategy}) to select solutions, o3 followed a much simpler approach. Specifically, we selected the top 50 solutions with the highest test-time compute from 1,024 samples per problem. Despite this streamlined method, o3 produced robust solutions capable of covering many, if not all, subtasks --- without the need for subtask-specific prompts, manual partitioning, or intricate submission strategies.

\begin{figure}[h!]
    \centering
    \includegraphics[width=0.85\textwidth]{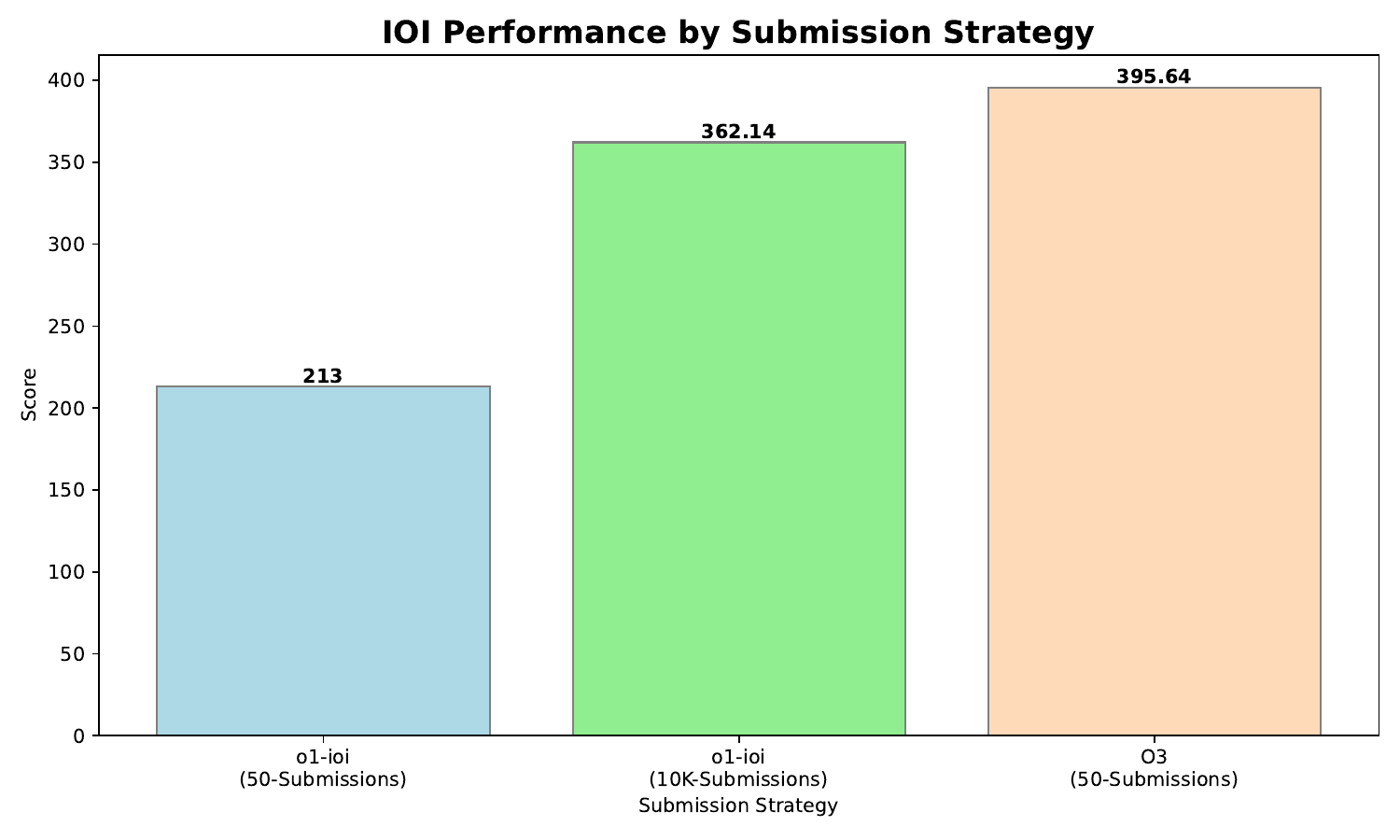}
    \caption{\textbf{IOI 2024 scores under different submission strategies.} Even without human-engineered heuristics or relaxed submission limits, o3 outperforms o1-ioi and surpasses the gold threshold with just 50 submissions.}
    \label{fig:o3_ioi_bar}
\end{figure}

\paragraph{Results.} \Cref{fig:o3_ioi_bar} presents the final scores. The IOI scoring system is subtask-based, with a maximum total of 600 points in the 2024 contest. The gold medal threshold was approximately 360 points. Key results include:

\begin{itemize}[leftmargin=15pt]
    \item o1-ioi scored 213 points with 50 submissions, improving to 362.14 points with 10K submissions, just above the gold medal cutoff.
    \item o3 achieved 395.64 points, surpassing the gold threshold even under the 50-submission limit.
\end{itemize}

These results demonstrate that o3 outperforms o1-ioi without relying on IOI-specific, hand-crafted test-time strategies. Instead, the sophisticated test-time techniques that emerged during o3 training, such as generating brute-force solutions to verify outputs, served as a more than adequate replacement and eliminated the need for the hand-engineered clustering and selection pipelines required by o1-ioi.

Overall, the IOI 2024 findings confirm that large-scale RL training alone can achieve state-of-the-art coding and reasoning performance. By independently learning to generate, evaluate, and refine solutions, o3 surpasses o1-ioi without dependence on domain-specific heuristics or clustering-based methods.

\section{Software Engineering Evaluations}
We have demonstrated how reasoning significantly enhances LLM performance in competitive programming, where solving complex algorithmic challenges requires deep logical thinking. However, we also sought to evaluate the impact of reasoning on real-world coding tasks. To this end, we tested our models on two datasets: the HackerRank Astra\footnote{\url{www.hackerrank.com/ai/astra}}  dataset and SWE-bench verified\footnote{\url{https://openai.com/index/introducing-swe-bench-verified/}}~\cite{jimenez2023swe,openai_swe_bench_verified}.

\subsection{HackerRank Astra}

The HackerRank Astra dataset is composed of 65 project-oriented coding challenges, each crafted to simulate real-world software development tasks. These challenges cover a range of frameworks, including React.js, Django, and Node.js, allowing for hands-on experience in building features and applications.

What sets this dataset apart is its focus on assessing problem-solving skills in complex, multi-file, long-context scenarios that mirror actual development environments. Unlike typical competitive programming datasets, HackerRank Astra does not provide public test cases, which prevents us from relying on hand-crafted test-time tactics. Evaluating performance with this dataset reveals whether reasoning abilities enhance success in algorithmic problem solving alone, or extend to more practical, industry-related coding tasks.

\begin{figure}[h!] \centering \includegraphics[width=0.85\textwidth]{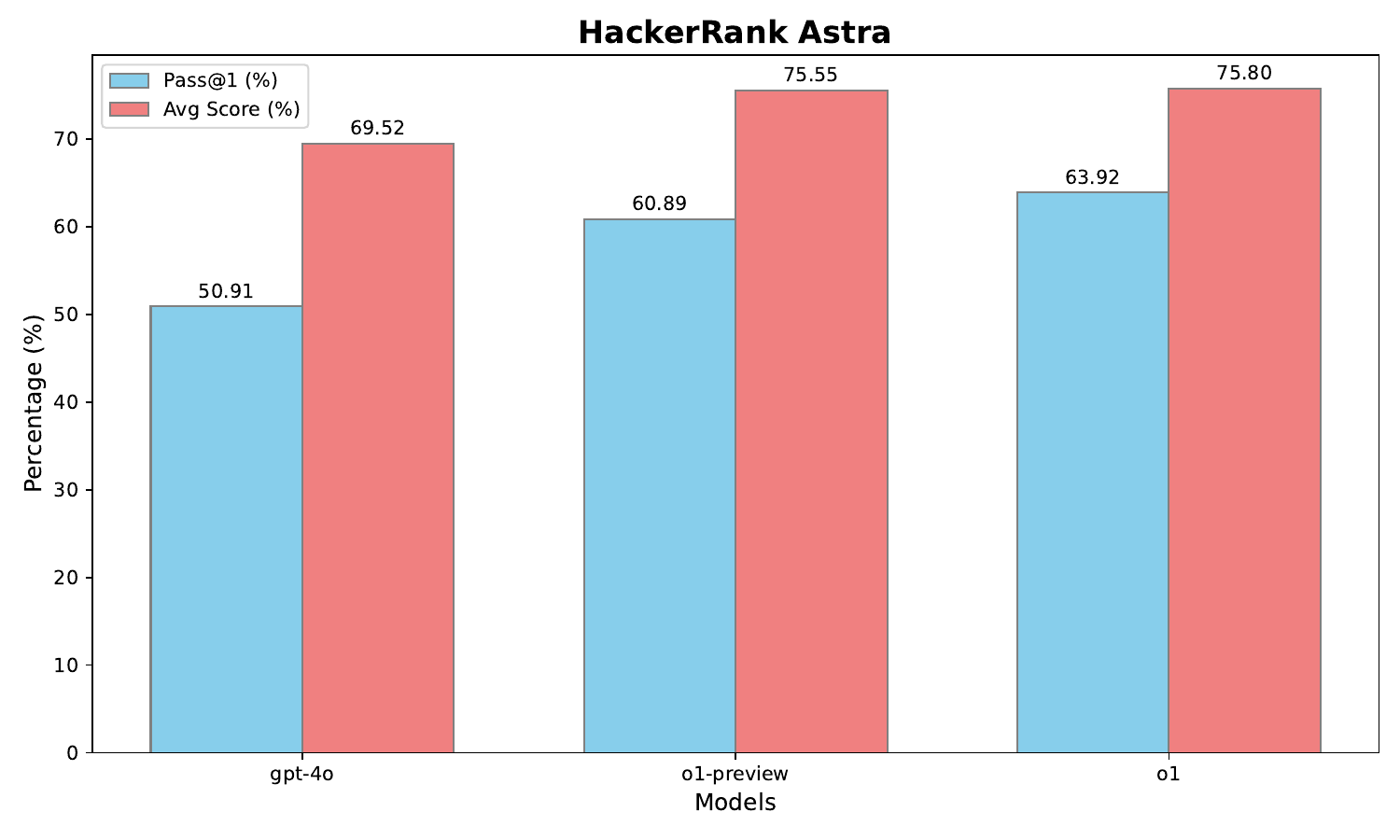} \caption{HackerRank Astra evaluation.} \label{fig:astra} \end{figure}

Figure~\ref{fig:astra} presents performance metrics such as pass@1 (the probability of successfully completing a task on the first attempt) and average scores (the mean proportion of test cases passed). The results illustrate the impact of chain-of-thought reasoning, with the o1-preview model achieving a 9.98\% improvement in pass@1 and a 6.03-point gain in average score compared to GPT-4o. Further fine-tuning through reinforcement learning enhances o1's performance, yielding a pass@1 of 63.92\% and an average score of 75.80\%—a 3.03\% increase in pass@1 over o1-preview. These metrics demonstrate o1's enhanced reasoning and adaptability, enabling it to address complex, industry-relevant software development tasks effectively.

\subsection{SWE-Bench Verified}
SWE-bench Verified is OpenAI's preparedness team's human-validated subset of SWE-bench that more reliably evaluates AI models’ ability to solve real-world software issues. This validated set of 500 tasks fixes certain issues with SWE-bench such as incorrect grading of correct solutions, under-specified problem statements, and overly specific unit tests. This helps ensure the benchmark accurately grades model capabilities.

To illustrate performance on this software task, we display the results presented in the o1 system card~\cite{jaech2024openai} as well as results from an early o3 checkpoint~\cite{openaio3}. Because o1-preview was not trained to use code execution or file editing tools, the best-performing open-source scaffold at the time of initial implementation, Agentless was used. Unlike for IOI, no specialized test-time strategies was used for SWE-Bench verified. All models are given 5 tries to generate a candidate patch. If the model fails after 5 attempts, it is considered an incorrect attempt. All evaluations are averaged over 3 trials. We do not penalize the model for system failures (e.g., container hangs or grading failures), and we retry these rollouts until we can record a valid attempt.

\begin{figure}[H] \centering \includegraphics[width=0.85\textwidth]{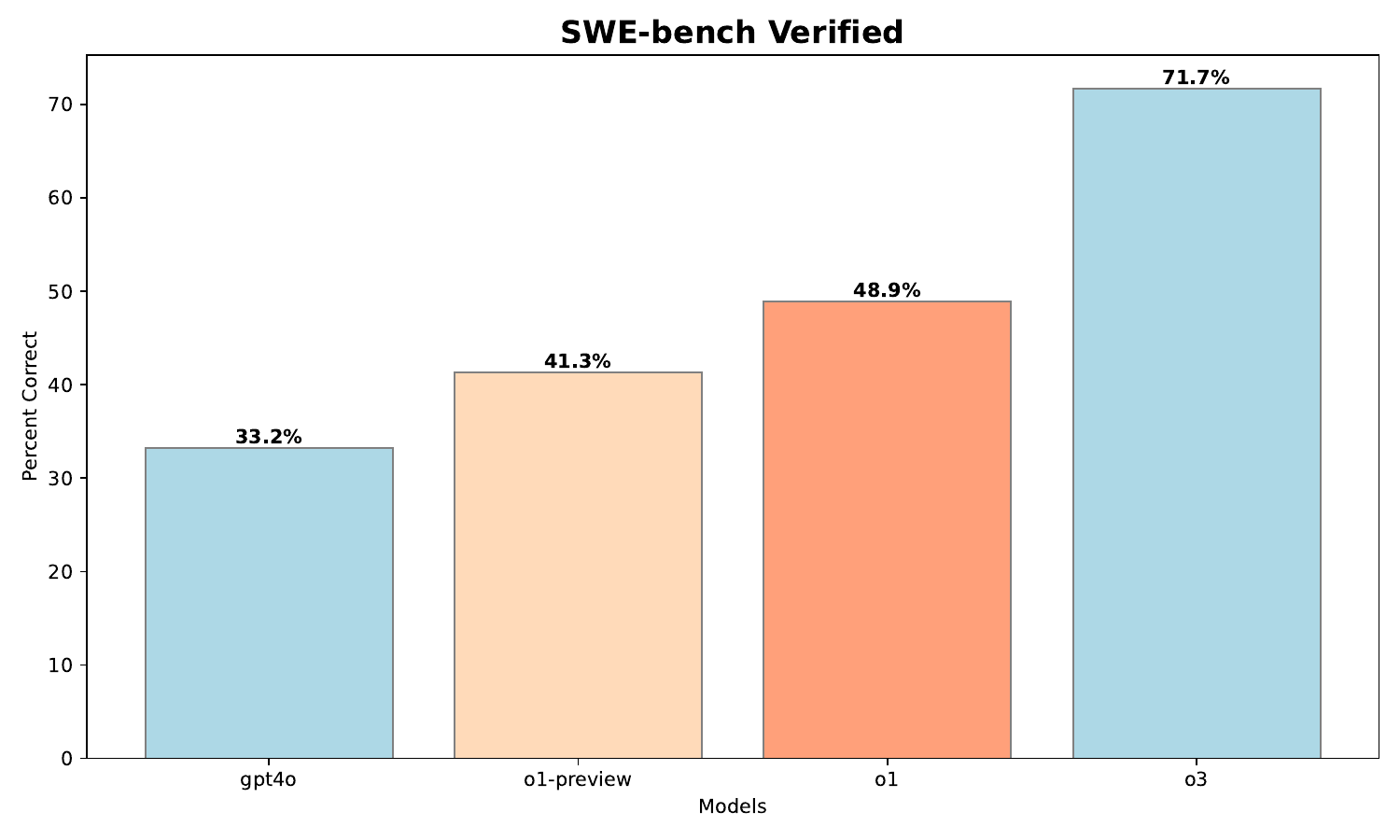} \caption{SWE-bench evaluation.} \label{fig:swebench} \end{figure}

As illustrated in Figure~\ref{fig:swebench}, o1-preview demonstrates an 8.1\% performance improvement on SWE-bench compared to gpt-4o, showcasing notable advancements in reasoning capabilities. With additional reinforcement learning compute applied during training, o1 achieves a further 8.6\% improvement. Notably, o3, which was trained with significantly greater compute resources than o1, delivers an impressive 22.8\% improvement over o1. These results underscore that enhanced reasoning skills extend beyond competitive programming challenges, proving their applicability to real-world tasks like software engineering.

\section{Conclusion}
Through the o-series large reasoning models, we demonstrate that chain-of-thought reasoning is a powerful strategy for improving performance in coding tasks, from competitive programming benchmarks such as \codeforces and IOI to complex software engineering challenges like SWE-bench and Astra. Our findings highlight that increasing reinforcement learning training compute, coupled with enhanced test-time compute, consistently boosts model performance to nearly match the best humans in the world. Given these results, we believe o-series large reasoning models will unlock many new use cases for AI in science, coding, math, and many other fields.
\appendix

\section{Authorship, credit attribution, and acknowledgments}
\label{app:authors}

\paragraph{Data Preparation:}
Borys Minaiev, Ignasi Clavera, Lorenz Kuhn, Nat McAleese, Oleg Mürk, Szymon Sidor

\paragraph{IOI Model Training:} Ahmed El-Kishky, Mostafa Rohaninejad

\paragraph{Sampling Infrastructure:}
Andre Saraiva, Hunter Lightman, Vineet Kosaraju, Wenda Zhou

\paragraph{Test-time Strategy:} Alexander Wei, Daniel Selsam, David Dohan, Francis Song, Ignasi Clavera, Max Schwarzer, Rhythm Garg, Rui Shu

\paragraph{Evaluation:} Andre Saraiva, Ignasi Clavera, Lorenz Kuhn, Nat McAleese

\paragraph{Leadership:}
Jakub Pachocki, Jerry Tworek, Lukasz Kaiser, Mark Chen

\paragraph{o3 Model Development} o3 contributors~\cite{openaio3}.

\paragraph{Acknowledgments:} We are grateful to the IOI committee for allowing us to enter our model, o1-ioi, in the 2024 International Olympiad in Informatics. We also extend our thanks to Wael Ewida, a member of the IOI technical committee, for hosting a portal that enabled us to submit our solutions under the same conditions as the contestants. Additionally, we appreciate the support of those who contributed to and maintained our sandboxed code execution, including Taylor Gordon, Oleg Boiko, John Rizzo, Paul Ashbourne, Leo Liu, Alexander Prokofiev, and Scottie Yan. We also extend our gratitude to Chris Orsinger and Michelle Fradin for their contributions to data efforts. Finally, we would like to express our sincere gratitude to everyone involved in the reinforcement learning reasoning efforts for o1 and o3, whose dedication and expertise were instrumental in advancing this work.

\section{Additional \codeforces Details}
\label{app:cf_details}
In order to compare our models to human competitive programmers, we simulate contests.
This section provides details of how the simulation is performed, how the overall score and ratings are calculated, as well as the per-contest results.

\subsection{Data}
For our test set we use ``Division 1'' contests from late 2023 and 2024, all of which occurred after the o3 training set data cut-off. 
As a redundant additional check, we used embedding search to confirm that the test problems have not been seen by the model during training.
We excluded one contest that contained an interactive problem for which grading was inconvenient, but otherwise included all post-cut-off Division 1 problems to which we had access at the time.
During training we used a validation set of primarily Division 2 problems; when that set indicated that performance was very strong we built and evaluated the Division 1 set presented here.

\subsection{Grading}
We run the complete set of tests for each problem, and have confirmed that our test environment closely matches the official \codeforces grading service, including by manually submitting solutions for the hardest problems to the official \codeforces graders.

Following AlphaCode \cite{leblond2023alphacode} we allow the model to make 10 independent submissions against the full test set and mark a problem as solved if any one of those 10 passes.
This is close to but not strictly the same as the human affordance, as human participants see only the results of the pre-tests during the competition.
However in Division 1 contests the pre-tests are typically ``strong'' (highly correlated with full tests), and in our results the number of failures before a passing submission is typically small (see \ref{tab:cf-details}).
We did not have access to labels for which test cases were pre-tests.

\subsection{Thinking Time}

Competitors receive a higher score for submitting their solutions faster. 
Because models can think in parallel and simultaneously attempt all problems, they have an innate advantage over humans. We elected to reduce this advantage in our primary results by estimating o3's score for each solved problem as the median of the scores of the human participants that solved that problem in the contest with the same number of failed attempts.

We could instead use the model's real thinking time to compute ratings.
o3 uses a learned scoring function for test-time ranking in addition to a chain of thought.
This process is perfectly parallel and true model submission times therefore depend on the number of available GPU during the contest.
On a very large cluster the time taken to pick the top-ranked solutions is (very slightly more than) the maximum over the thinking times for each candidate submission.
Using this maximum parallelism assumption and the sequential o3 sampling speed would result in a higher estimated rating than presented here.
We note that because sequential test-time compute has grown rapidly since the early language models, it was not guaranteed that models would solve problems quickly compared to humans, but in practice o3 does.

\subsection{Estimated Rating}

The \codeforces rating system is described by the creator in three blog posts \cite{cf_rating_1, cf_rating_2, cf_rating_3}.
Ratings are similar to the Elo system and satisfy the property that if competitor $A$ has rating $R_A$ and competitor $B$ has rating $R_B$ then the probability that $A$ ranks better than $B$ any final contest standings is estimated as $$\frac 1 {10^{\frac {R_B - R_A} {400}}}$$

To find the model rating we first calculate the rank of the model in each of the test contest from the total contest score (described above) and then directly maximize the likelihood of the observed rankings and human ratings with respect to the model rating using the equation above.
We average to ensure that contests with more participants are not over-weighted.

We validated that this recovers known human ratings based on their contest performance and also gives similar values to linearly predicting participant rating from their average solve rate.

\subsection{Percentile performance}

\begin{figure}[h!]
    \centering
    \includegraphics[width=0.75\textwidth]{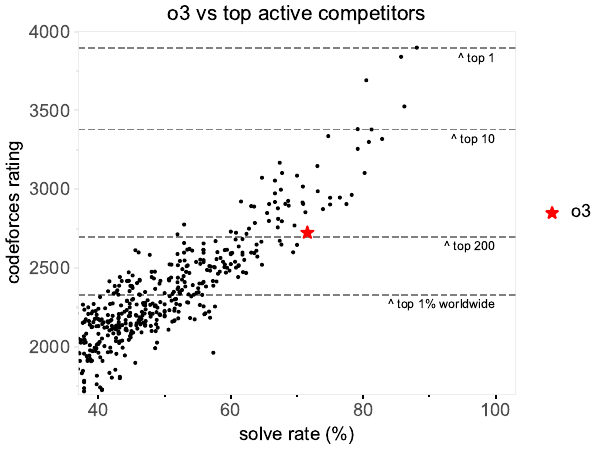}
    \caption{o3 would place among the best human competitive programmers in the world. Here we show the average solve rate and current rating for participants that entered at least 8 of our 12 unseen test contests. Horizontal lines show performance thresholds from the global \codeforces leaderboard of active competitors. The very best humans still solve more problems than AI, for now.}
    \label{fig:o3-vs-best-humans}
\end{figure}

Codeforces maintains a global leaderboard of active participants, and an estimated rating can be used to compare to that group. 
We can also directly compare the solve rate of o3 in our test contests to the other participants in those contests.
Figure \ref{fig:o3-vs-best-humans} shows both these comparisons.
Each point is a person that competed in at least 8 of the test contests.
We show their average solve rate over contests that they entered against their rating, as well as the rating thresholds for key performance levels.
The very best human competitors remain much stronger than o3, with solve rates in excess of 85\%, but both ratings and solve rates indicate that o3 would rank among the top 200 active participants worldwide.

\subsection{Per Problem Breakdown}

{\small
\begin{longtable}{lrrrrr}
\caption{
We estimate our \codeforces rating from simulated contest participation.
Here we show a detailed breakdown of o3 performance per-problem.
}%
\label{tab:cf-details}\\

\toprule
\textbf{problem} &
\shortstack{\textbf{problem}\\\textbf{rating}} &
\shortstack{\textbf{pass@1}\\\textbf{(no ranking)}} &
\shortstack{\textbf{pass@10}\\\textbf{(no ranking)}} &
\shortstack{\textbf{\# failed}\\\textbf{submissions}} &
\shortstack{\textbf{pass@10}\\\textbf{(ranking 1162)}}\\
\midrule
\endfirsthead

\toprule
\textbf{problem} &
\shortstack{\textbf{problem}\\\textbf{rating}} &
\shortstack{\textbf{pass@1}\\\textbf{(no ranking)}} &
\shortstack{\textbf{pass@10}\\\textbf{(no ranking)}} &
\shortstack{\textbf{\# failed}\\\textbf{submissions}} &
\shortstack{\textbf{pass@10}\\\textbf{(ranking 1162)}}\\
\midrule
\endhead

\midrule
\multicolumn{6}{r}{\textit{Continued on the next page}}\\
\endfoot

\bottomrule
\endlastfoot
\multicolumn{6}{l}{\textbf{\shortstack[l]{\textbf{Contest 1909 - 23/Dec/23 - Pinely Round 3 (Div. 1 + Div. 2)}\\\textbf{ score: 7,220}}}}\\
1909 A & 800 & 1156 / 1162 & 1.00 & 0 & solved\\
1909 B & 1200 & 1066 / 1162 & 1.00 & 0 & solved\\
1909 C & 1400 & 1075 / 1162 & 1.00 & 0 & solved\\
1909 D & 1900 & 1099 / 1162 & 1.00 & 0 & solved\\
1909 E & 2400 & 703 / 1162 & 1.00 & 0 & solved\\
1909 F1 & 2200 & 57 / 1162 & 0.40 & 0 & solved\\
1909 F2 & 2500 & 0 / 1162 & 0.00 & 0 & not solved\\
1909 G & 3000 & 3 / 1162 & 0.03 & 0 & not solved\\
1909 H & 3500 & 0 / 1162 & 0.00 & 0 & not solved\\
1909 I & 1900 & 0 / 1162 & 0.00 & 0 & not solved\\
\multicolumn{6}{l}{\textbf{\shortstack[l]{\textbf{Contest 1916 - 30/Dec/23 - Good Bye 2023}\\\textbf{ score: 8,920}}}}\\
1916 A & 800 & 1157 / 1162 & 1.00 & 0 & solved\\
1916 B & 1000 & 1133 / 1162 & 1.00 & 0 & solved\\
1916 C & 1200 & 1145 / 1162 & 1.00 & 0 & solved\\
1916 D & 1700 & 483 / 1162 & 1.00 & 0 & solved\\
1916 E & 2300 & 6 / 1162 & 0.05 & 0 & solved\\
1916 F & 2900 & 369 / 1162 & 0.98 & 2 & solved\\
1916 G & 3500 & 0 / 1162 & 0.00 & 0 & not solved\\
1916 H1 & 2700 & 1059 / 1162 & 1.00 & 0 & solved\\
1916 H2 & 2700 & 1045 / 1162 & 1.00 & 0 & solved\\
\multicolumn{6}{l}{\textbf{\shortstack[l]{\textbf{Contest 1919 - 06/Jan/24 - Hello 2024}\\\textbf{ score: 6,214}}}}\\
1919 A & 800 & 1161 / 1162 & 1.00 & 0 & solved\\
1919 B & 800 & 1141 / 1162 & 1.00 & 0 & solved\\
1919 C & 1400 & 499 / 1162 & 1.00 & 0 & solved\\
1919 D & 2100 & 25 / 1162 & 0.20 & 2 & solved\\
1919 E & 2600 & 6 / 1162 & 0.05 & 1 & solved\\
1919 F1 & 2300 & 1090 / 1162 & 1.00 & 0 & solved\\
1919 F2 & 2800 & 227 / 1162 & 0.89 & 0 & solved\\
1919 G & 3500 & 0 / 1162 & 0.00 & 0 & not solved\\
1919 H & 2000 & 0 / 1162 & 0.00 & 0 & not solved\\
\multicolumn{6}{l}{\textbf{\shortstack[l]{\textbf{Contest 1942 - 30/Mar/24 - CodeTON Round 8 (Div. 1 + Div. 2, Rated, Prizes!)}\\\textbf{ score: 8,701}}}}\\
1942 A & 800 & 1157 / 1162 & 1.00 & 0 & solved\\
1942 B & 1100 & 1157 / 1162 & 1.00 & 0 & solved\\
1942 C1 & 1300 & 999 / 1162 & 1.00 & 0 & solved\\
1942 C2 & 1700 & 525 / 1162 & 1.00 & 1 & solved\\
1942 D & 2100 & 1061 / 1162 & 1.00 & 0 & solved\\
1942 E & 2300 & 347 / 1162 & 0.97 & 0 & solved\\
1942 F & 2700 & 0 / 1162 & 0.00 & 0 & not solved\\
1942 G & 2800 & 239 / 1162 & 0.90 & 0 & solved\\
1942 H & 3500 & 0 / 1162 & 0.00 & 0 & not solved\\
\multicolumn{6}{l}{\textbf{\shortstack[l]{\textbf{Contest 1943 - 16/Mar/24 - Codeforces Round 934 (Div. 1)}\\\textbf{ score: 3,427}}}}\\
1943 A & 1300 & 116 / 1162 & 0.65 & 0 & solved\\
1943 B & 2000 & 1 / 1162 & 0.01 & 0 & not solved\\
1943 C & 2300 & 160 / 1162 & 0.77 & 0 & solved\\
1943 D1 & 2400 & 848 / 1162 & 1.00 & 0 & solved\\
1943 D2 & 2800 & 14 / 1162 & 0.11 & 0 & solved\\
1943 E1 & 2900 & 0 / 1162 & 0.00 & 0 & not solved\\
1943 E2 & 3300 & 0 / 1162 & 0.00 & 0 & not solved\\
1943 F & 3500 & 0 / 1162 & 0.00 & 0 & not solved\\
\multicolumn{6}{l}{\textbf{\shortstack[l]{\textbf{Contest 1951 - 06/Apr/24 - Codeforces Global Round 25}\\\textbf{ score: 9,396}}}}\\
1951 A & 900 & 1157 / 1162 & 1.00 & 0 & solved\\
1951 B & 1200 & 1150 / 1162 & 1.00 & 0 & solved\\
1951 C & 1400 & 1155 / 1162 & 1.00 & 0 & solved\\
1951 D & 2000 & 875 / 1162 & 1.00 & 0 & solved\\
1951 E & 2000 & 1009 / 1162 & 1.00 & 0 & solved\\
1951 F & 2500 & 53 / 1162 & 0.37 & 0 & solved\\
1951 G & 3100 & 34 / 1162 & 0.26 & 0 & solved\\
1951 H & 3200 & 1 / 1162 & 0.01 & 0 & not solved\\
1951 I & 3200 & 0 / 1162 & 0.00 & 0 & not solved\\
\multicolumn{6}{l}{\textbf{\shortstack[l]{\textbf{Contest 1965 - 27/Apr/24 - Codeforces Round 941 (Div. 1)}\\\textbf{ score: 3,891}}}}\\
1965 A & 1400 & 1143 / 1162 & 1.00 & 0 & solved\\
1965 B & 1800 & 1064 / 1162 & 1.00 & 0 & solved\\
1965 C & 2300 & 313 / 1162 & 0.96 & 0 & solved\\
1965 D & 2900 & 690 / 1162 & 1.00 & 0 & solved\\
1965 E & 3100 & 0 / 1162 & 0.00 & 0 & not solved\\
1965 F & 3300 & 0 / 1162 & 0.00 & 0 & not solved\\
\multicolumn{6}{l}{\textbf{\shortstack[l]{\textbf{Contest 1967 - 30/Apr/24 - Codeforces Round 942 (Div. 1)}\\\textbf{ score: 3,871}}}}\\
1967 A & 1400 & 1088 / 1162 & 1.00 & 0 & solved\\
1967 B1 & 1400 & 1154 / 1162 & 1.00 & 0 & solved\\
1967 B2 & 2200 & 1149 / 1162 & 1.00 & 0 & solved\\
1967 C & 2300 & 1116 / 1162 & 1.00 & 0 & solved\\
1967 D & 2800 & 9 / 1162 & 0.08 & 0 & solved\\
1967 E1 & 3100 & 0 / 1162 & 0.00 & 0 & not solved\\
1967 E2 & 3500 & 0 / 1162 & 0.00 & 0 & not solved\\
1967 F & 3200 & 0 / 1162 & 0.00 & 0 & not solved\\
\multicolumn{6}{l}{\textbf{\shortstack[l]{\textbf{Contest 1975 - 25/May/24 - Codeforces Round 947 (Div. 1 + Div. 2)}\\\textbf{ score: 5,959}}}}\\
1975 A & 800 & 1161 / 1162 & 1.00 & 0 & solved\\
1975 B & 1000 & 1091 / 1162 & 1.00 & 0 & solved\\
1975 C & 1200 & 492 / 1162 & 1.00 & 0 & solved\\
1975 D & 1700 & 9 / 1162 & 0.08 & 3 & solved\\
1975 E & 2100 & 80 / 1162 & 0.51 & 1 & solved\\
1975 F & 2600 & 12 / 1162 & 0.10 & 0 & solved\\
1975 G & 3000 & 0 / 1162 & 0.00 & 0 & not solved\\
1975 H & 3500 & 0 / 1162 & 0.00 & 0 & not solved\\
1975 I & 3500 & 0 / 1162 & 0.00 & 0 & not solved\\
\multicolumn{6}{l}{\textbf{\shortstack[l]{\textbf{Contest 1984 - 09/Jun/24 - Codeforces Global Round 26}\\\textbf{ score: 12,255}}}}\\
1984 A & 800 & 1161 / 1162 & 1.00 & 0 & solved\\
1984 B & 1100 & 1158 / 1162 & 1.00 & 0 & solved\\
1984 C1 & 1300 & 914 / 1162 & 1.00 & 0 & solved\\
1984 C2 & 1700 & 768 / 1162 & 1.00 & 0 & solved\\
1984 D & 2000 & 193 / 1162 & 0.84 & 1 & solved\\
1984 E & 2400 & 849 / 1162 & 1.00 & 1 & solved\\
1984 F & 2500 & 918 / 1162 & 1.00 & 0 & solved\\
1984 G & 3200 & 0 / 1162 & 0.00 & 0 & not solved\\
1984 H & 3300 & 138 / 1162 & 0.72 & 3 & solved\\
\multicolumn{6}{l}{\textbf{\shortstack[l]{\textbf{Contest 2002 - 11/Aug/24 - EPIC IoT Round August 2024 (Div. 1 + Div. 2)}\\\textbf{ score: 8,981}}}}\\
2002 A & 800 & 1161 / 1162 & 1.00 & 0 & solved\\
2002 B & 1000 & 1152 / 1162 & 1.00 & 0 & solved\\
2002 C & 1200 & 1096 / 1162 & 1.00 & 0 & solved\\
2002 D1 & 1900 & 1067 / 1162 & 1.00 & 0 & solved\\
2002 D2 & 2300 & 805 / 1162 & 1.00 & 0 & solved\\
2002 E & 2300 & 232 / 1162 & 0.89 & 0 & solved\\
2002 F1 & 2600 & 12 / 1162 & 0.10 & 8 & solved\\
2002 F2 & 2800 & 7 / 1162 & 0.06 & 0 & not solved\\
2002 G & 3400 & 143 / 1162 & 0.73 & 1 & solved\\
2002 H & 3500 & 0 / 1162 & 0.00 & 0 & not solved\\
\multicolumn{6}{l}{\textbf{\shortstack[l]{\textbf{Contest 2006 - 30/Aug/24 - Codeforces Round 969 (Div. 1)}\\\textbf{ score: 2,757}}}}\\
2006 A & 1700 & 945 / 1162 & 1.00 & 0 & solved\\
2006 B & 1800 & 1149 / 1162 & 1.00 & 0 & solved\\
2006 C & 2300 & 811 / 1162 & 1.00 & 0 & solved\\
2006 D & 2600 & 2 / 1162 & 0.02 & 0 & not solved\\
2006 E & 3100 & 4 / 1162 & 0.03 & 0 & not solved\\
2006 F & 3500 & 0 / 1162 & 0.00 & 0 & not solved\\
\end{longtable}
}

\section{IOI Submissions}
\label{appendix:ioi_samples}
This section presents the solutions generated by o1-ioi during the 2024 International Olympiad in Informatics.
\input{samples/nile}
\input{samples/message}
\input{samples/tree}
\input{samples/hieroglyphs}
\input{samples/mosaic}
\input{samples/sphinx}

\bibliographystyle{plain}
\bibliography{main}
\end{document}

%% file: samples/nile.tex
\subsection{Nile}
All 100 possible points for Nile were scored in a single submission.

\begin{lstlisting}[language=C++]
#include "nile.h"
#include <bits/stdc++.h>
using namespace std;
 
struct UF{
    vector<int> parent, sizee, minIndex;
    vector<array<long long, 2>> singledPat; // singledPat[0] = min d[i] for i mod 2 = 0 in block, singledPat[1] for i mod 2 = 1 in block
    vector<long long> singledBr; // singled bridging
    vector<long long> singled; // singled in block
};
 
long long sumSingledGlobal = 0;
 
int findRoot(UF &u, int x){
    if(u.parent[x] == x) return x;
    u.parent[x] = findRoot(u, u.parent[x]);
    return u.parent[x];
}
 
void updateSingled(UF &u, int r){
    if(u.sizee[r] %
        u.singled[r] = 0;
    } else {
        long long val = min(u.singledPat[r][u.minIndex[r] %
        if(val == LLONG_MAX) val = 0; // if no such node found, singled = 0?
        u.singled[r] = val;
    }
}
 
void unionNodes(UF &u, int a, int b){
    a = findRoot(u, a);
    b = findRoot(u, b);
    if(a == b) return; // already in the same component
 
    sumSingledGlobal -= u.singled[a];
    sumSingledGlobal -= u.singled[b];
 
    // union by size
    if(u.sizee[a] < u.sizee[b]) swap(a, b);
    u.parent[b] = a;
 
    int newSize = u.sizee[a] + u.sizee[b];
    int newMinIndex = min(u.minIndex[a], u.minIndex[b]);
 
    array<long long, 2> newSingledPat;
    newSingledPat[0] = min(u.singledPat[a][0], u.singledPat[b][0]);
    newSingledPat[1] = min(u.singledPat[a][1], u.singledPat[b][1]);
 
    long long newSingledBr = min(u.singledBr[a], u.singledBr[b]);
 
    u.sizee[a] = newSize;
    u.minIndex[a] = newMinIndex;
    u.singledPat[a] = newSingledPat;
    u.singledBr[a] = newSingledBr;
 
    updateSingled(u, a);
 
    sumSingledGlobal += u.singled[a];
}
 
void addBridging(UF &u, int x, long long val){
    int r = findRoot(u, x);
    sumSingledGlobal -= u.singled[r];
    if(u.sizee[r] %
        u.singledBr[r] = min(u.singledBr[r], val);
        updateSingled(u, r);
    } else {
        // block size is even, singled = 0 anyway, bridging won't matter because singledBr is for odd sized block.
        // but bridging might matter if block size becomes odd in future merges, so store singledBr as well.
        u.singledBr[r] = min(u.singledBr[r], val);
        // singled is 0 anyway
    }
    sumSingledGlobal += u.singled[r];
}
 
std::vector<long long> calculate_costs(
    std::vector<int> W, std::vector<int> A,
    std::vector<int> B, std::vector<int> E){
 
    int N = W.size();
    int Q = E.size();
 
    vector<int> idx(N); iota(idx.begin(), idx.end(), 0);
    sort(idx.begin(), idx.end(), [&](int a, int b){return W[a] < W[b];});
 
    vector<long long> sortedW(N), d(N);
    for(int i = 0; i < N; i++) {
        sortedW[i] = W[idx[i]];
        d[i] = (long long)A[idx[i]] - (long long)B[idx[i]];
    }
 
    long long sumB = 0;
    for(int i = 0; i < N; i++) sumB += B[idx[i]];
 
    struct Edge{ long long diff; int idx;};
    vector<Edge> edges;
    edges.reserve(N-1);
    for(int i = 0; i < N-1; i++) {
        edges.push_back({sortedW[i+1] - sortedW[i], i});
    }
    sort(edges.begin(), edges.end(), [&](const Edge &a, const Edge &b){return a.diff < b.diff;});
 
    vector<pair<long long, int>> bridging;
    bridging.reserve(N-2);
    for(int i = 1; i < N-1; i++) {
        long long val = sortedW[i+1] - sortedW[i-1];
        bridging.push_back({val, i});
    }
    sort(bridging.begin(), bridging.end(), [&](auto &a, auto &b){return a.first < b.first;});
 
    vector<pair<long long, int>> queriesVec;
    queriesVec.reserve(Q);
    for(int i = 0; i < Q; i++) queriesVec.push_back({(long long)E[i], i});
    sort(queriesVec.begin(), queriesVec.end(), [&](auto &a, auto &b){return a.first < b.first;});
 
    UF u;
    u.parent.resize(N);
    u.sizee.resize(N);
    u.minIndex.resize(N);
    u.singledPat.resize(N);
    u.singledBr.resize(N);
    u.singled.resize(N);
 
    for(int i = 0; i < N; i++) {
        u.parent[i] = i;
        u.sizee[i] = 1;
        u.minIndex[i] = i;
        u.singledPat[i][0] = (i %
        u.singledPat[i][1] = (i %
        u.singledBr[i] = LLONG_MAX;
        u.singled[i] = d[i]; // singled = d[i] because block size 1 is odd
    }
 
    sumSingledGlobal = 0;
    for(int i = 0; i < N; i++) {
        sumSingledGlobal += u.singled[i];
    }
 
    vector<long long> result(Q);
 
    int posE = 0; // index in edges
    int posB = 0; // index in bridging
 
    for(int i = 0; i < Q; i++) {
        long long D = queriesVec[i].first;
        int qidx = queriesVec[i].second;
 
        while(posE < (int)edges.size() && edges[posE].diff <= D) {
            unionNodes(u, edges[posE].idx, edges[posE].idx+1);
            posE++;
        }
 
        while(posB < (int)bridging.size() && bridging[posB].first <= D) {
            int x = bridging[posB].second;
            addBridging(u, x, d[x]);
            posB++;
        }
 
        long long costVal = sumB + sumSingledGlobal;
        result[qidx] = costVal;
    }
 
    return result;
}
\end{lstlisting}

%% file: samples/message.tex
\subsection{Message}
This solution achieved a score of 79.64 out of 100, with full marks awarded for subtask 1 and partial credit received for subtask 2.

\begin{lstlisting}[language=C++]
#include "message.h"
#include <bits/stdc++.h>
using namespace std;
 
extern std::vector<bool> send_packet(std::vector<bool> A);
 
void send_message(std::vector<bool> M, std::vector<bool> C)
{
    vector<int> safe_cols;
    vector<int> sabotage_cols;
    for (int i = 0; i < 31; i++) {
        if (!C[i]) safe_cols.push_back(i);
        else sabotage_cols.push_back(i);
    }
 
    // Step 1: send the first 4 packets with distinct sequences.
    vector<vector<bool>> first4results;
    first4results.reserve(4);
    for (int p = 0; p < 4; p++) {
        vector<bool> A(31, false);
        for (int i = 0; i < 16; i++) {
            int col = safe_cols[i];
            int bit = (i >> p) & 1; // p-th bit of i, i is the index in safe subset
            A[col] = bit;
        }
        for (int col : sabotage_cols) {
            A[col] = false;
        }
        std::vector<bool> B = send_packet(A);
        first4results.push_back(B);
    }
 
    // Now identify known safe column from the results of these 4 calls.
    vector<string> colSequences(31, string(4, '0'));
    for (int c = 0; c < 31; c++) {
        for (int r = 0; r < 4; r++) {
            colSequences[c][r] = first4results[r][c] ? '1' : '0';
        }
    }
 
    map<string, vector<int>> seqToCols;
    for (int c = 0; c < 31; c++) {
        seqToCols[colSequences[c]].push_back(c);
    }
    int knownSafeCol = -1;
    for (auto &kv : seqToCols) {
        if (kv.second.size() == 1) {
            int col = kv.second[0];
            // check if col is in safe subset. Because known safe column must be a safe column.
            if (find(safe_cols.begin(), safe_cols.end(), col) != safe_cols.end()) {
                // also check if the sequence is one of the distinct sequences for safe subset. The index in the safe subset is i, the sequence is 4 bits representing i.
                string seq = kv.first;
                int val = 0;
                for (int i = 0; i < 4; i++) {
                    if (seq[i] == '1') val |= (1 << i);
                }
                // check if val is < 16.
                if (val < 16) {
                    knownSafeCol = col;
                    break;
                }
            }
        }
    }
    if (knownSafeCol == -1) {
        // If not found, pick a safe column that has a distinct sequence anyway.
        // But for correctness, we require knownSafeCol to be found. If not found, our method has a problem.
    }
 
    // Step 2: store sabotage subset's bits in the known safe column in next 31 calls. The other 15 safe columns store message bits if available.
    int S = M.size(); // message length.
    int offset = 0; // offset in M.
    for (int i = 0; i < 31; i++) {
        vector<bool> A(31, false);
        int sabotageBit = C[i];
        A[knownSafeCol] = sabotageBit; // store sabotage subset's bit i in known safe column.
        for (int c : safe_cols) {
            if (c == knownSafeCol) continue; // skip known safe column because we used it for sabotage subset.
            int col = c;
            if (offset < S) {
                A[col] = M[offset];
                offset++;
            } else {
                A[col] = false;
            }
        }
        for (int sc : sabotage_cols) {
            A[sc] = false;
        }
        std::vector<bool> B = send_packet(A);
        // We do not store B because we do not need to adapt further. Actually, we can store B if we want to adapt but the solution does not require further adaptation.
    }
 
    // Step 3: store message length in the next 11 calls in the known safe column. The other 15 safe columns store more message bits if available.
    int L = S; // length in integer form.
    vector<int> lengthBinary(11, 0);
    for (int i = 0; i < 11; i++) {
        lengthBinary[i] = (L >> i) & 1;
    }
 
    for (int i = 0; i < 11; i++) {
        vector<bool> A(31, false);
        int bit = lengthBinary[i];
        A[knownSafeCol] = bit; // store length bit i in known safe column.
        for (int c : safe_cols) {
            if (c == knownSafeCol) continue; // skip known safe column.
            int col = c;
            if (offset < S) {
                A[col] = M[offset];
                offset++;
            } else {
                A[col] = false;
            }
        }
        for (int sc : sabotage_cols) {
            A[sc] = false;
        }
        std::vector<bool> B = send_packet(A);
    }
 
    // Step 4: store the rest of the message in subsequent calls using all 16 safe columns.
    int messageLeft = S - offset; // how many bits are left to store.
    int fullColumns = 16; // number of safe columns.
    int messageRows = (messageLeft + fullColumns - 1) / fullColumns; // how many rows needed.
 
    for (int r = 0; r < messageRows; r++) {
        vector<bool> A(31, false);
        for (int i = 0; i < fullColumns && offset + i < S; i++) {
            int col = safe_cols[i];
            A[col] = M[offset + i];
        }
        offset += fullColumns;
        for (int sc : sabotage_cols) {
            A[sc] = false;
        }
        std::vector<bool> B = send_packet(A);
    }
 
}
 
std::vector<bool> receive_message(std::vector<std::vector<bool>> R)
{
    int n = R.size();
    vector<bool> result; // message to return.
    if (n < 4) {
        return result;
    }
 
    // Step 1: from the first 4 rows in R, find the known safe column.
    vector<string> colSequences(31, string(4, '0'));
    for (int c = 0; c < 31; c++) {
        for (int r = 0; r < 4 && r < n; r++) {
            if (r < n) {
                colSequences[c][r] = R[r][c] ? '1' : '0';
            }
        }
    }
 
    map<string, vector<int>> seqToCols;
    for (int c = 0; c < 31; c++) {
        seqToCols[colSequences[c]].push_back(c);
    }
    int knownSafeCol = -1;
    for (auto &kv : seqToCols) {
        if (kv.second.size() == 1) {
            int col = kv.second[0];
            // check if the sequence is one of the binary representations from 0..15 because safe subset's sequences are distinct and in that range.
            string seq = kv.first;
            int val = 0;
            for (int i = 0; i < 4; i++) {
                if (seq[i] == '1') val |= (1 << i);
            }
            if (val < 16) {
                knownSafeCol = col;
                break;
            }
        }
    }
    if (knownSafeCol == -1) {
        // If none found, that means sabotage overcame the method or we made an error.
    }
 
    // Step 2: from next 31 rows in known safe column, read sabotage subset.
    vector<bool> sabotageSubset(31, false);
    int sabotageStart = 4;
    for (int i = 0; i < 31 && sabotageStart + i < n; i++) {
        sabotageSubset[i] = R[sabotageStart + i][knownSafeCol];
    }
 
    // Step 3: from next 11 rows in known safe column, read length.
    vector<bool> lengthBits(11, false);
    int lengthStart = sabotageStart + 31;
    for (int i = 0; i < 11 && lengthStart + i < n; i++) {
        lengthBits[i] = R[lengthStart + i][knownSafeCol];
    }
    int messageLength = 0;
    for (int i = 0; i < 11; i++) {
        messageLength |= (lengthBits[i] << i);
    }
 
    // Step 4: identify safe columns from sabotage subset.
    vector<int> safe_cols;
    for (int i = 0; i < 31; i++) {
        if (!sabotageSubset[i]) safe_cols.push_back(i);
    }
 
    // Step 5: read message from the next (31 + 11) rows' other safe columns and from subsequent rows using all safe columns.
    vector<bool> message;
    int offset = 0; // offset in message.
    int sabotageRowStart = 4;
    int sabotageRows = 31;
    for (int r = sabotageRowStart; r < sabotageRowStart + sabotageRows && r < n; r++) {
        for (int c : safe_cols) {
            if (c == knownSafeCol) continue; // skip known safe column.
            if (offset < messageLength) {
                message.push_back(R[r][c]);
                offset++;
            }
        }
    }
 
    int lengthRowStart = sabotageRowStart + sabotageRows; // 4 + 31 = 35.
    int lengthRows = 11;
    for (int r = lengthRowStart; r < lengthRowStart + lengthRows && r < n; r++) {
        for (int c : safe_cols) {
            if (c == knownSafeCol) continue; // skip known safe column.
            if (offset < messageLength) {
                message.push_back(R[r][c]);
                offset++;
            }
        }
    }
 
    int restRowStart = lengthRowStart + lengthRows; // 4 + 31 + 11 = 46.
    int fullColumns = 16; // number of safe columns.
    int messageLeft = messageLength - offset;
 
    int rowsForMessage = (messageLeft + fullColumns - 1) / fullColumns; // number of rows needed.
    for (int r = restRowStart; r < restRowStart + rowsForMessage && r < n; r++) {
        for (int i = 0; i < fullColumns && offset < messageLength; i++) {
            int col = safe_cols[i];
            message.push_back(R[r][col]);
            offset++;
        }
    }
 
    // The message is now in vector message. If offset < messageLength, we did not get all bits. but we should.
    return message;
}
\end{lstlisting}

%% file: samples/tree.tex
\subsection{Tree}
A total of 30 points were scored on Tree across two separate submissions.

\subsubsection{Submission 1}
The first submission achieved a score of 17 out of 100, with points earned from subtasks 1 and 4.

\begin{lstlisting}[language=C++]
#include "tree.h"
#include <bits/stdc++.h>
using namespace std;
 
static int NGlobal;
static vector<int> parentGlobal;
static vector<int> weightGlobal;
static vector<long long> deltaGlobal;
static vector<int> leafCountGlobal;
static vector<int> sortedByLeafCount; // We'll store node indices sorted by leaf count.
static vector<long long> prefixDelta, prefixDeltaTimesLeafCount; // prefix sums in sorted order by leaf count.
static long long sumAllDelta;
static vector<vector<int>> childrenGlobal;
 
int computeLeafCountRec(int v){
    if (childrenGlobal[v].empty()){ // leaf
        leafCountGlobal[v] = 1;
        return 1;
    } else {
        int sum = 0;
        for (int c : childrenGlobal[v]) sum += computeLeafCountRec(c);
        leafCountGlobal[v] = sum;
        return sum;
    }
}
 
void init(std::vector<int> P, std::vector<int> W){
    int N = (int)P.size();
    NGlobal = N;
    parentGlobal = P;
    weightGlobal = W;
    if (N == 0) return;
 
    childrenGlobal.assign(N, {});
    for (int i=1; i < N; i++) childrenGlobal[parentGlobal[i]].push_back(i);
 
    leafCountGlobal.resize(N);
    computeLeafCountRec(0);
 
    vector<int> deg(N,0);
    for (int i=1; i < N; i++) deg[parentGlobal[i]]++;
    deltaGlobal.resize(N);
    for (int i=0; i < N; i++){
        if (parentGlobal[i] == -1){ // root
            if (deg[i] == 0) { // leaf root
                deltaGlobal[i] = weightGlobal[i];
            } else { // internal root
                deltaGlobal[i] = - (long long)weightGlobal[i];
            }
        } else {
            if (deg[i] == 0) { // leaf
                deltaGlobal[i] = (long long)weightGlobal[parentGlobal[i]] + (long long)weightGlobal[i];
            } else { // internal
                deltaGlobal[i] = (long long)weightGlobal[parentGlobal[i]] - (long long)weightGlobal[i];
            }
        }
    }
 
    sortedByLeafCount.resize(N);
    iota(sortedByLeafCount.begin(), sortedByLeafCount.end(), 0);
    sort(sortedByLeafCount.begin(), sortedByLeafCount.end(), [](int a, int b){
        return leafCountGlobal[a] < leafCountGlobal[b];
    });
    prefixDelta.resize(N,0);
    prefixDeltaTimesLeafCount.resize(N,0);
    sumAllDelta = 0;
    for (int i=0; i < N; i++) sumAllDelta += deltaGlobal[i];
    long long runningDeltaTimesLeaf = 0;
    long long runningDelta = 0;
    for (int i=0; i < N; i++){
        int idx = sortedByLeafCount[i];
        long long valDeltaTimesLeaf = deltaGlobal[idx] * (long long)leafCountGlobal[idx];
        runningDeltaTimesLeaf += valDeltaTimesLeaf;
        runningDelta += deltaGlobal[idx];
        prefixDeltaTimesLeafCount[i] = runningDeltaTimesLeaf;
        prefixDelta[i] = runningDelta;
    }
}
 
long long query(int L, int R){
    if (NGlobal == 0) return 0; // no nodes
    long long a = R / L; // floor of R/L.
 
    int left = -1;
    int right = NGlobal;
    while (right - left > 1){
        int mid = (left + right)/2;
        int node = sortedByLeafCount[mid];
        if ((long long)leafCountGlobal[node] <= a) left = mid; else right = mid;
    }
    int idx = left; // idx is the last index where b[i] <= a.
 
    long long sumA = 0; // sum of delta[i]*b[i] for nodes with b[i] <= a.
    if (idx >= 0) {
        sumA = prefixDeltaTimesLeafCount[idx]; // sum_{i: b[i] <= a} delta[i]*b[i]
    }
 
    long long sumPrefixDeltaWhereLessOrEqual = 0;
    if (idx >= 0) sumPrefixDeltaWhereLessOrEqual = prefixDelta[idx]; else sumPrefixDeltaWhereLessOrEqual = 0;
    long long sumDeltaWhereBGreater = sumAllDelta - sumPrefixDeltaWhereLessOrEqual;
 
    long long LLL = (long long)L;
    long long RRR = (long long)R;
 
    long long partA = LLL * sumA;
    long long partB = RRR * sumDeltaWhereBGreater;
 
    long long cost = partA + partB;
 
    return cost;
}
\end{lstlisting}

\subsubsection{Submission 2}
Submission 2 achieved 13 of 100 points on subtask 2.
\begin{lstlisting}[language=C++]
#include "tree.h"
#include <bits/stdc++.h>
using namespace std;
struct Pt {
    long long x;
    long long y;
};
struct PWL {
    long long L, R;
    vector<Pt> pts; // sorted by x
};
int NGlobal;
int PGlobal[2005];
int wGlobal[2005];
vector<int> childrenGlobal[2005];
PWL GGlobal[2005];
 
// define the helper functions as static.
 
static long long evalPWL(const PWL &f, long long x) {
    if (f.pts.empty()) return LLONG_MAX/2; // no function
    if (x <= f.pts.front().x) {
        return f.pts.front().y; // domain out, but we prefer front.
    }
    if (x >= f.pts.back().x) {
        return f.pts.back().y; // domain out, but prefer back.
    }
 
    int left = 0, right = f.pts.size()-1;
    while (left+1 < right) {
        int mid = (left + right) / 2;
        if (f.pts[mid].x == x) return f.pts[mid].y;
        if (f.pts[mid].x < x) left = mid; else right = mid;
    }
    if (f.pts[left].x <= x && x <= f.pts[right].x) {
        long long x1 = f.pts[left].x;
        long long y1 = f.pts[left].y;
        long long x2 = f.pts[right].x;
        long long y2 = f.pts[right].y;
        if (x1 == x2) return y1;
        long double slope = (long double)(y2 - y1)/(long double)(x2 - x1);
        long double valf = (long double) y1 + slope*(x - x1);
        long long vali = (long long) llround(valf);
        return vali;
    } else {
        return LLONG_MAX/2;
    }
}
struct ChildState {
    int idx; // child index in local array
    long long x; // current x in child's domain
    int seg; // segment in child's function
    long double slope;
    long long nextBound;
};
 
static PWL combineChildren(const vector<PWL> &childF, long long L, long long R) {
    int d = childF.size();
    if (d == 0) {
        PWL f; f.L = L; f.R = R; f.pts = {{L, 0}, {R, 0}}; return f;
    }
    long long sumDomainL = (long long)d * L;
    long long sumDomainR = (long long)d * R;
 
    vector<ChildState> cs(d);
    priority_queue<pair<long double, int>, vector<pair<long double, int>>, greater<pair<long double, int>>> pq;
 
    for (int i = 0; i < d; i++) {
        cs[i].idx = i;
        cs[i].x = L;
        int seg = 0; int n = childF[i].pts.size();
        while (seg < n-1 && childF[i].pts[seg+1].x <= L) seg++;
        cs[i].seg = seg;
        long double slope = 0;
        if (seg+1 < n) {
            long long dx = childF[i].pts[seg+1].x - childF[i].pts[seg].x;
            long long dy = childF[i].pts[seg+1].y - childF[i].pts[seg].y;
            slope = dx!=0?(long double)dy/dx:0;
        } else {
            slope = 0;
        }
        cs[i].slope = slope;
        if (seg+1 < n) cs[i].nextBound = min((long long)childF[i].pts[seg+1].x, R);
        else cs[i].nextBound = R;
        pq.push({slope, i});
    }
 
    long long sum = sumDomainL;
    long long cost = 0;
    for (int i = 0; i < d; i++) {
        cost += evalPWL(childF[i], L);
    }
 
    vector<Pt> fpts;
    fpts.push_back({sum, cost});
 
    while (!pq.empty() && sum < sumDomainR) {
        auto top = pq.top();
        long double slope = top.first;
        vector<int> group;
        while (!pq.empty() && abs(pq.top().first - slope) < 1e-9) {
            group.push_back(pq.top().second);
            pq.pop();
        }
        if (group.empty()) break;
 
        long long inc = LLONG_MAX;
        for (int idx : group) {
            long long newInc = cs[idx].nextBound - cs[idx].x;
            if (newInc < inc) inc = newInc;
        }
        if (inc <= 0) {
            for (int idx : group) {
                pq.push({cs[idx].slope, idx});
            }
            break;
        }
        long long feasibleInc = inc;
        if (sum + (long long)group.size() * inc > sumDomainR) {
            feasibleInc = (sumDomainR - sum) / (long long) group.size();
        }
        if (feasibleInc <= 0) {
            for (int idx : group) {
                pq.push({cs[idx].slope, idx});
            }
            break;
        }
 
        sum = sum + group.size() * feasibleInc;
        long double dcost = slope * feasibleInc * group.size();
        cost = (long long) llround((long double)cost + dcost);
 
        fpts.push_back({sum, cost});
 
        for (int idx : group) {
            cs[idx].x += feasibleInc;
            if (cs[idx].x == cs[idx].nextBound) {
                const PWL &g = childF[idx];
                int seg = cs[idx].seg;
                if (cs[idx].x == R) {
                    cs[idx].slope = 1e9; // saturate
                    cs[idx].nextBound = R;
                } else {
                    seg++;
                    cs[idx].seg = seg;
                    int n = g.pts.size();
                    if (seg+1 < n) {
                        long long dx = g.pts[seg+1].x - g.pts[seg].x;
                        long long dy = g.pts[seg+1].y - g.pts[seg].y;
                        long double slopeC = dx!=0?(long double)dy/dx:0;
                        cs[idx].slope = slopeC;
                        cs[idx].nextBound = min((long long)g.pts[seg+1].x, R);
                    } else {
                        cs[idx].slope = 0;
                        cs[idx].nextBound = R;
                    }
                }
            }
        }
        for (int idx : group) {
            if (cs[idx].x < R) {
                pq.push({cs[idx].slope, idx});
            }
        }
    }
 
    vector<Pt> unify;
    unify.push_back(fpts[0]);
    for (int i = 1; i < fpts.size(); i++) {
        if (fpts[i].x == unify.back().x) {
            unify.back().y = min(unify.back().y, fpts[i].y);
        } else unify.push_back(fpts[i]);
    }
 
    vector<Pt> final;
    if (!unify.empty()) final.push_back(unify[0]);
    for (int i = 1; i < unify.size()-1; i++) {
        long long x1 = final.back().x;
        long long y1 = final.back().y;
        long long x2 = unify[i].x;
        long long y2 = unify[i].y;
        long long x3 = unify[i+1].x;
        long long y3 = unify[i+1].y;
        long double slope1 = (x2!=x1)? (long double)(y2-y1)/(x2-x1) : 1e9;
        long double slope2 = (x3!=x2)? (long double)(y3-y2)/(x3-x2) : 1e9;
        if (abs(slope1 - slope2) < 1e-9) {
            // unify linear, skip unify[i]
        } else {
            final.push_back(unify[i]);
        }
    }
    if (!unify.empty()) final.push_back(unify.back());
 
    PWL result;
    result.L = sumDomainL; result.R = sumDomainR;
    result.pts = final;
 
    return result;
}
 
static PWL parentFormula(const PWL &f, long long wVal, long long L, long long R) {
    vector<long long> candVal;
    candVal.push_back(L);
    candVal.push_back(R);
    for (auto &p : f.pts) {
        if (p.x >= L && p.x <= R) candVal.push_back(p.x);
    }
    if (f.L >= L && f.L <= R) candVal.push_back(f.L);
    if (f.R >= L && f.R <= R) candVal.push_back(f.R);
 
    for (int i = 0; i < (int)f.pts.size()-1; i++) {
        long long sx = f.pts[i].x;
        long long sy = f.pts[i+1].x;
        long long dx = sy - sx;
        long long dy = f.pts[i+1].y - f.pts[i].y;
        long double a = dx!=0? (long double)dy/dx:0; // slope in s domain
        if (a <= wVal && a >= -wVal) {
            long long inL = max(sx, (long long)L);
            long long inR = min(sy, (long long)R);
            if (inL <= inR) {
                candVal.push_back(inL);
                candVal.push_back(inR);
            }
        }
    }
 
    sort(candVal.begin(), candVal.end());
    candVal.erase(unique(candVal.begin(), candVal.end()), candVal.end());
 
    vector<Pt> pts;
    for (auto val : candVal) {
        long long bestCost = LLONG_MAX/2;
        if (val >= f.L && val <= f.R) {
            int left = 0, right = f.pts.size()-1;
            while (left+1 < right) {
                int mid = (left + right) / 2;
                if (f.pts[mid].x <= val) left = mid; else right = mid;
            }
            if (f.pts[left].x <= val && val <= f.pts[right].x) {
                long long x1 = f.pts[left].x;
                long long y1 = f.pts[left].y;
                long long x2 = f.pts[right].x;
                long long y2 = f.pts[right].y;
                long double slope = (x2!=x1? (long double)(y2 - y1)/(long double)(x2 - x1) : 0);
                if (slope <= wVal && slope >= -wVal && val >= x1 && val <= x2) {
                    long double costf = (long double) y1 + slope * (long double)(val - x1);
                    long long c = (long long) llround(costf);
                    bestCost = min(bestCost, c);
                }
            }
        }
        for (auto &p : f.pts) {
            long long s = p.x;
            long long c = p.y + wVal * llabs(val - s);
            if (c < bestCost) bestCost = c;
        }
        pts.push_back({val, bestCost});
    }
 
    sort(pts.begin(), pts.end(), [](const Pt &a, const Pt &b){return a.x < b.x;});
 
    vector<Pt> unify;
    unify.push_back(pts[0]);
    for (int i = 1; i < pts.size(); i++) {
        if (pts[i].x == unify.back().x) {
            unify.back().y = min(unify.back().y, pts[i].y);
        } else unify.push_back(pts[i]);
    }
 
    vector<Pt> final;
    if (!unify.empty()) final.push_back(unify[0]);
    for (int i = 1; i < unify.size()-1; i++) {
        long long x1 = final.back().x;
        long long y1 = final.back().y;
        long long x2 = unify[i].x;
        long long y2 = unify[i].y;
        long long x3 = unify[i+1].x;
        long long y3 = unify[i+1].y;
        long double slope1 = x2!=x1? (long double)(y2-y1)/(x2-x1) : 1e9;
        long double slope2 = x3!=x2? (long double)(y3-y2)/(x3-x2) : 1e9;
        if (abs(slope1 - slope2) < 1e-9) {
            // unify linear, skip unify[i]
        } else {
            final.push_back(unify[i]);
        }
    }
    if (!unify.empty()) final.push_back(unify.back());
 
    PWL result;
    result.L = L; result.R = R;
    result.pts = final;
 
    return result;
}
 
static void computeG(int node, long long L, long long R) {
    for (int c : childrenGlobal[node]) {
        computeG(c, L, R);
    }
    if (childrenGlobal[node].empty()) {
        // leaf
        PWL result;
        result.L = L; result.R = R;
        if (L <= 0 && 0 <= R) {
            long long valL = wGlobal[node] * llabs(L);
            long long val0 = wGlobal[node] * 0; // 0
            long long valR = wGlobal[node] * llabs(R);
            result.pts.clear();
            result.pts.push_back({L, valL});
            result.pts.push_back({0, val0});
            result.pts.push_back({R, valR});
        } else {
            long long valL = wGlobal[node] * llabs(L);
            long long valR = wGlobal[node] * llabs(R);
            result.pts.clear();
            result.pts.push_back({L, valL});
            result.pts.push_back({R, valR});
        }
        GGlobal[node] = result;
    } else {
        vector<PWL> childF;
        for (int c : childrenGlobal[node]) {
            childF.push_back(GGlobal[c]);
        }
        PWL f = combineChildren(childF, L, R);
        PWL p = parentFormula(f, wGlobal[node], L, R);
        GGlobal[node] = p;
    }
}
 
void init(std::vector<int> P, std::vector<int> W) {
    NGlobal = W.size();
    for (int i = 0; i < NGlobal; i++) {
        PGlobal[i] = P[i];
        wGlobal[i] = W[i];
    }
    for (int i = 0; i < NGlobal; i++) childrenGlobal[i].clear();
    for (int i = 1; i < NGlobal; i++) {
        childrenGlobal[P[i]].push_back(i);
    }
}
 
long long query(int L, int R) {
    computeG(0, L, R);
    long long ans = LLONG_MAX;
    for (auto &p : GGlobal[0].pts) {
        if (p.y < ans) ans = p.y;
    }
    return ans;
}
\end{lstlisting}

%% file: samples/hieroglyphs.tex
\subsection{Hieroglyphs}
A total of 44 points was scored on Hieroglyphs across two separate submissions.

\subsubsection{Submission 1}
In the first submission, a score of 34 out of 100 points was achieved, distributed across subtasks 1, 2, and 4.

\begin{lstlisting}[language=C++]
#include "hieroglyphs.h"
#include <bits/stdc++.h>
using namespace std;
std::vector<int> ucs(std::vector<int> A, std::vector<int> B)
{
    int N = (int) A.size();
    int M = (int) B.size();
    if (N == 0 || M == 0) {
        std::vector<int> ret;
        return ret;
    }
    const int MAX_VAL = 200001;
 
    static int cntA[200002], cntB[200002];
    for (int i = 0; i < 200002; i++) {
        cntA[i] = 0;
        cntB[i] = 0;
    }
 
    for (int i = 0; i < N; i++) cntA[A[i]]++;
    for (int i = 0; i < M; i++) cntB[B[i]]++;
 
    static int cVal[200002];
    for (int i = 0; i < 200002; i++) {
        if (cntA[i] > 0 && cntB[i] > 0) {
            cVal[i] = min(cntA[i], cntB[i]);
        } else {
            cVal[i] = 0;
        }
    }
 
    long long sumC = 0;
    for (int x = 0; x < 200002; x++) {
        sumC += cVal[x];
    }
 
    if (sumC == 0) {
        std::vector<int> ret;
        return ret;
    }
 
    // Precompute earliest and latest occurrences in A and B for all x.
    static int earliestA[200002], latestA[200002], earliestB[200002], latestB[200002];
    for (int i = 0; i < 200002; i++) {
        earliestA[i] = -1;
        latestA[i] = -1;
        earliestB[i] = -1;
        latestB[i] = -1;
    }
 
    for (int i = 0; i < N; i++) {
        int letter = A[i];
        if (earliestA[letter] == -1) earliestA[letter] = i;
        latestA[letter] = i;
    }
    for (int i = 0; i < M; i++) {
        int letter = B[i];
        if (earliestB[letter] == -1) earliestB[letter] = i;
        latestB[letter] = i;
    }
 
    // Now intervals for cVal[x] = 1.
    struct Interval {
        int letter;
        int eA, lA, eB, lB;
    };
 
    vector<Interval> intervals;
    for (int x = 0; x < 200002; x++) {
        if (cVal[x] == 1) {
            Interval I;
            I.letter = x;
            I.eA = earliestA[x]; I.lA = latestA[x];
            I.eB = earliestB[x]; I.lB = latestB[x];
            intervals.push_back(I);
        }
    }
 
    // sort intervals by eA.
    std::sort(intervals.begin(), intervals.end(), [](const Interval &a, const Interval &b){
        return a.eA < b.eA;
    });
 
    // Balanced tree keyed by eB, store lB as well.
    vector<int> segmentTree(4*(M+5), -1);
    auto update = [&](int idx, int start, int end, int pos, int val, auto &&f) -> void {
        if (start == end) {
            segmentTree[idx] = val;
            return;
        }
        int mid = (start+end)/2;
        if (pos <= mid) f(idx*2, start, mid, pos, val, f);
        else f(idx*2+1, mid+1, end, pos, val, f);
        segmentTree[idx] = max(segmentTree[idx*2], segmentTree[idx*2+1]);
    };
    auto query = [&](int idx, int start, int end, int l, int r, auto &&f) -> int {
        if (r < start || end < l) return -1;
        if (l <= start && end <= r) {
            return segmentTree[idx];
        }
        int mid = (start+end)/2;
        int leftVal = f(idx*2, start, mid, l, r, f);
        int rightVal = f(idx*2+1, mid+1, end, l, r, f);
        return max(leftVal, rightVal);
    };
 
    vector<Interval> intervalsSortedByLA = intervals;
    sort(intervalsSortedByLA.begin(), intervalsSortedByLA.end(), [](const Interval &a, const Interval &b){
        return a.lA < b.lA;
    });
 
    int j = 0;
    for (int i = 0; i < intervals.size(); i++) {
        auto &x = intervals[i];
        int eAx = x.eA;
        // remove intervals from data structure where lA[y] < eAx.
        while (j < intervalsSortedByLA.size() && intervalsSortedByLA[j].lA < eAx) {
            auto &y = intervalsSortedByLA[j];
            // remove y from segment tree keyed by eB[y].
            update(1, 0, M-1, y.eB, -1, update);
            j++;
        }
 
        // query in B dimension: find if there's an interval y with eB[y] <= lB[x] and lB[y] >= eB[x].
        int minB = x.eB; // eB[x]
        int maxB = x.lB; // lB[x]
        if (minB > maxB) {
            // If eB[x] > lB[x], no intersection possible.
        } else {
            // query in the segment tree range [0, maxB] to find the maximum lB[y].
            int maxVal = query(1, 0, M-1, 0, maxB, query);
            if (maxVal >= minB) {
                // found intersection with a letter y where cVal[y] = 1.
                std::vector<int> ret;
                ret.push_back(-1);
                return ret;
            }
        }
 
        // add x to data structure keyed by eB[x].
        update(1, 0, M-1, x.eB, x.lB, update);
    }
 
    // If no intersection found among cVal=1 intervals, proceed with chain method.
 
    // The chain method code.
    static vector<int> posAarr[200002];
    static vector<int> posBarr[200002];
    for (int i = 0; i < 200002; i++) {
        posAarr[i].clear();
        posBarr[i].clear();
    }
 
    for (int i = 0; i < N; i++) {
        if (cVal[A[i]] > 0) posAarr[A[i]].push_back(i);
    }
    for (int i = 0; i < M; i++) {
        if (cVal[B[i]] > 0) posBarr[B[i]].push_back(i);
    }
 
    struct QItem {
        int letter;
        int pos;
    };
 
    struct QComp {
        bool operator()(const QItem &a, const QItem &b) const {
            if (a.pos == b.pos) return a.letter > b.letter;
            return a.pos > b.pos;
        }
    };
 
    priority_queue<QItem, vector<QItem>, QComp> pq;
 
    static int posAarrIdx[200002];
    static int occInAarr[200002];
 
    for (int x = 0; x < 200002; x++) {
        posAarrIdx[x] = 0;
        occInAarr[x] = 0;
        if (cVal[x] > 0 && posAarr[x].size() > 0) {
            QItem item;
            item.letter = x;
            item.pos = posAarr[x][0];
            pq.push(item);
        }
    }
 
    static int TAlpha[200002];
    for (int x = 0; x < 200002; x++) {
        if (cVal[x] > 0) TAlpha[x] = cntB[x] - cVal[x]; else TAlpha[x] = 0;
    }
 
    static int freqAlpha[200002]; // freq in B up to posB0.
    for (int x = 0; x < 200002; x++) freqAlpha[x] = 0;
 
    int posB0Alpha = -1;
    {
        int i = 0;
        for (; i < M; i++) {
            int letter = B[i];
            if (cVal[letter] > 0) {
                freqAlpha[letter]++;
                if (freqAlpha[letter] > TAlpha[letter]) {
                    freqAlpha[letter]--;
                    break; // cause posB0Alpha = i-1
                }
            }
        }
        posB0Alpha = i-1;
        if (i == M) posB0Alpha = M-1;
    }
 
    vector<int> U;
    U.reserve(sumC);
 
    int posB = -1;
    int usedCount = 0;
 
    auto updatePosB0Alpha = [&](int letter, auto &freqAlpha, auto &TAlpha, int &posB0Alpha, int M) {
        int i = posB0Alpha + 1;
        while (i < M) {
            int l = B[i];
            if (cVal[l] > 0) {
                freqAlpha[l]++;
                if (freqAlpha[l] > TAlpha[l]) {
                    freqAlpha[l]--;
                    break;
                }
            }
            i++;
        }
        posB0Alpha = i-1;
    };
 
    while (!pq.empty() && usedCount < sumC) {
        QItem item = pq.top();
        pq.pop();
 
        int letter = item.letter;
        int posAVal = posAarr[letter][posAarrIdx[letter]];
 
        auto &barr = posBarr[letter];
        int idx = (int) (std::lower_bound(barr.begin(), barr.end(), posB + 1) - barr.begin());
        if (idx == (int) barr.size()) {
            posAarrIdx[letter]++;
            if (posAarrIdx[letter] < posAarr[letter].size()) {
                QItem newItem;
                newItem.letter = letter;
                newItem.pos = posAarr[letter][posAarrIdx[letter]];
                pq.push(newItem);
            }
            continue;
        } else {
            int posBVal = barr[idx];
 
            TAlpha[letter]++;
            int oldPosB0Alpha = posB0Alpha;
            vector<pair<int,int>> freqChanges;
 
            int startIndex = posB0Alpha + 1;
            int i = startIndex;
            while (i < M) {
                int l = B[i];
                if (cVal[l] > 0) {
                    freqAlpha[l]++;
                    freqChanges.push_back({l, i});
                    if (freqAlpha[l] > TAlpha[l]) {
                        freqAlpha[l]--;
                        freqChanges.pop_back();
                        break;
                    }
                }
                i++;
            }
            int newPosB0Alpha = i-1;
 
            if (posBVal <= newPosB0Alpha) {
                U.push_back(letter);
                usedCount++;
                posB = posBVal;
 
                occInAarr[letter]++;
                posAarrIdx[letter]++;
                if (occInAarr[letter] < cVal[letter] && posAarrIdx[letter] < posAarr[letter].size()) {
                    QItem newItem;
                    newItem.letter = letter;
                    newItem.pos = posAarr[letter][posAarrIdx[letter]];
                    pq.push(newItem);
                }
 
                posB0Alpha = newPosB0Alpha;
 
            } else {
                TAlpha[letter]--;
                for (auto &fc : freqChanges) {
                    freqAlpha[fc.first]--;
                }
                posB0Alpha = oldPosB0Alpha;
 
                posAarrIdx[letter]++;
                if (posAarrIdx[letter] < posAarr[letter].size()) {
                    QItem newItem;
                    newItem.letter = letter;
                    newItem.pos = posAarr[letter][posAarrIdx[letter]];
                    pq.push(newItem);
                }
                continue;
            }
 
        }
    }
 
    if (usedCount < sumC) {
        std::vector<int> ret;
        ret.push_back(-1);
        return ret;
    }
 
    return U;
}
\end{lstlisting}

\subsubsection{Submission 2}
In the second submission, the model scored 10 points on subtask 3.

\begin{lstlisting}[language=C++]
#include "hieroglyphs.h"
#include <bits/stdc++.h>
using namespace std;
 
std::vector<int> ucs(std::vector<int> A, std::vector<int> B)
{
    int N = (int)A.size();
    int M = (int)B.size();
 
    int a0 = 0, a1 = 0;
    for (int i = 0; i < N; i++) {
        if (A[i] == 0) a0++;
        else if (A[i] == 1) a1++;
    }
    int b0 = 0, b1 = 0;
    for (int i = 0; i < M; i++) {
        if (B[i] == 0) b0++;
        else if (B[i] == 1) b1++;
    }
 
    int Z = min(a0, b0);
    int O = min(a1, b1);
 
    // If Z=0 and O=0, universal subsequence is the empty sequence.
    if (Z == 0 && O == 0) {
        return {};
    }
 
    vector<int> zeroPosA; zeroPosA.reserve(a0);
    for (int i = 0; i < N; i++) if (A[i] == 0) zeroPosA.push_back(i);
    vector<int> zeroPosB; zeroPosB.reserve(b0);
    for (int i = 0; i < M; i++) if (B[i] == 0) zeroPosB.push_back(i);
 
    vector<int> prefixOnesA(N+1, 0);
    for (int i = 1; i <= N; i++) prefixOnesA[i] = prefixOnesA[i-1] + (A[i-1] == 1);
    vector<int> prefixOnesB(M+1, 0);
    for (int i = 1; i <= M; i++) prefixOnesB[i] = prefixOnesB[i-1] + (B[i-1] == 1);
 
    auto getOnesAfterZero = [&](vector<int> & zeroPos, vector<int> & prefixOnes, int totalOnes, int z) {
        if (z == 0) return totalOnes;
        else {
            if (z-1 >= (int)zeroPos.size()) return 0; // safety
            int pos = zeroPos[z-1];
            int onesUsed = prefixOnes[pos+1];
            int onesAfter = totalOnes - onesUsed;
            return onesAfter;
        }
    };
 
    vector<int> fAval(Z+1), fBval(Z+1);
    for (int z = 0; z <= Z; z++) {
        fAval[z] = getOnesAfterZero(zeroPosA, prefixOnesA, a1, z);
        fBval[z] = getOnesAfterZero(zeroPosB, prefixOnesB, b1, z);
    }
 
    vector<int> cZ(Z+1);
    for (int z = 0; z <= Z; z++) cZ[z] = min(fAval[z], fBval[z]);
 
    vector<int> onePosA; onePosA.reserve(a1);
    for (int i = 0; i < N; i++) if (A[i] == 1) onePosA.push_back(i);
    vector<int> onePosB; onePosB.reserve(b1);
    for (int i = 0; i < M; i++) if (B[i] == 1) onePosB.push_back(i);
 
    vector<int> prefixZerosA(N+1, 0);
    for (int i = 1; i <= N; i++) prefixZerosA[i] = prefixZerosA[i-1] + (A[i-1] == 0);
    vector<int> prefixZerosB(M+1, 0);
    for (int i = 1; i <= M; i++) prefixZerosB[i] = prefixZerosB[i-1] + (B[i-1] == 0);
 
    auto getZerosAfterOne = [&](vector<int> & onePos, vector<int> & prefixZeros, int totalZeros, int w) {
        if (w == 0) return totalZeros;
        else {
            if (w-1 >= (int)onePos.size()) return 0; // safety
            int pos = onePos[w-1];
            int zerosUsed = prefixZeros[pos+1];
            int zerosAfter = totalZeros - zerosUsed;
            return zerosAfter;
        }
    };
 
    vector<int> gAval(O+1), gBval(O+1);
    for (int w = 0; w <= O; w++) {
        gAval[w] = getZerosAfterOne(onePosA, prefixZerosA, a0, w);
        gBval[w] = getZerosAfterOne(onePosB, prefixZerosB, b0, w);
    }
 
    vector<int> cO(O+1);
    for (int w = 0; w <= O; w++) cO[w] = min(gAval[w], gBval[w]);
 
    vector<int> U;
    U.reserve(Z+O);
 
    int zUsed = 0;
    int oUsed = 0;
 
    auto canPlaceZero = [&](int zUsed, int oUsed) {
        if (zUsed >= Z) return false;
        int c = cZ[zUsed+1]; // constraint for the next zero in U.
        int onesLeft = O - oUsed;
        return onesLeft >= c;
    };
 
    auto canPlaceOne = [&](int zUsed, int oUsed) {
        if (oUsed >= O) return false;
        int c = cO[oUsed+1]; // constraint for the next one in U.
        int zerosLeft = Z - zUsed;
        return zerosLeft >= c;
    };
 
    for (int i = 0; i < Z+O; i++) {
        bool placeZ = canPlaceZero(zUsed, oUsed);
        bool placeO = canPlaceOne(zUsed, oUsed);
        if (!placeZ && !placeO) {
            return {-1};
        } else if (placeZ && !placeO) {
            U.push_back(0);
            zUsed++;
        } else if (!placeZ && placeO) {
            U.push_back(1);
            oUsed++;
        } else {
            int cZc = cZ[zUsed+1];
            int cOc = cO[oUsed+1];
            bool canPickZero = false;
            bool canPickOne = false;
 
            {
                int newZUsed = zUsed + 1;
                int newOUsed = oUsed;
                int onesLeft = O - newOUsed;
                int zerosLeft = Z - newZUsed;
                bool feasible = true;
                if (Z > newZUsed) { // check cZ in range [newZUsed+1..Z]
                    int maxZConstraint = cZ[newZUsed+1];
                    if (onesLeft < maxZConstraint) feasible = false;
                }
                if (O > newOUsed) { // check cO in range [newOUsed+1..O]
                    int maxOConstraint = cO[newOUsed+1];
                    if (zerosLeft < maxOConstraint) feasible = false;
                }
                canPickZero = feasible;
            }
 
            {
                int newZUsed = zUsed;
                int newOUsed = oUsed + 1;
                int onesLeft = O - newOUsed;
                int zerosLeft = Z - newZUsed;
                bool feasible = true;
                if (Z > newZUsed) {
                    int maxZConstraint = cZ[newZUsed+1];
                    if (onesLeft < maxZConstraint) feasible = false;
                }
                if (O > newOUsed) {
                    int maxOConstraint = cO[newOUsed+1];
                    if (zerosLeft < maxOConstraint) feasible = false;
                }
                canPickOne = feasible;
            }
 
            if (canPickZero && !canPickOne) {
                U.push_back(0);
                zUsed++;
            } else if (!canPickZero && canPickOne) {
                U.push_back(1);
                oUsed++;
            } else if (canPickZero && canPickOne) {
                if (cZc > cOc) {
                    U.push_back(0);
                    zUsed++;
                } else if (cOc > cZc) {
                    U.push_back(1);
                    oUsed++;
                } else {
                    if ((Z - zUsed) < (O - oUsed)) {
                        U.push_back(0);
                        zUsed++;
                    } else {
                        U.push_back(1);
                        oUsed++;
                    }
                }
            } else {
                return {-1};
            }
        }
    }
 
    // Validate U with constraints.
    vector<int> zeroPositions, onePositions;
    zeroPositions.reserve(Z);
    onePositions.reserve(O);
    for (int i = 0; i < (int)U.size(); i++) {
        if (U[i] == 0) zeroPositions.push_back(i);
        else onePositions.push_back(i);
    }
 
    for (int z = 0; z <= Z; z++) {
        int pos = (z == 0) ? -1 : zeroPositions[z-1];
        int onesRight = (int)onePositions.size() - (int)(std::lower_bound(onePositions.begin(), onePositions.end(), pos+1) - onePositions.begin());
        if (onesRight < cZ[z]) {
            return {-1};
        }
    }
 
    for (int w = 0; w <= O; w++) {
        int pos = (w == 0) ? -1 : onePositions[w-1];
        int zerosRight = (int)zeroPositions.size() - (int)(std::lower_bound(zeroPositions.begin(), zeroPositions.end(), pos+1) - zeroPositions.begin());
        if (zerosRight < cO[w]) {
            return {-1};
        }
    }
 
    auto isSubsequence = [&](const std::vector<int> &C, const std::vector<int> &X) {
        int idx = 0;
        for (int i = 0; i < (int)X.size() && idx < (int)C.size(); i++) {
            if (X[i] == C[idx]) idx++;
        }
        return idx == (int)C.size();
    };
 
    if (!isSubsequence(U, A) || !isSubsequence(U, B)) {
        return {-1};
    }
 
    return U;
}
\end{lstlisting}

%% file: samples/mosaic.tex
\subsection{Mosaic}
A total of 42 points were scored on Mosaic across two separate submissions.

\subsubsection{Submission 1}
The first submission achieved a score of 22 out of 100, with points distributed across subtasks 1, 2, and 4.

\begin{lstlisting}[language=C++]
#include "mosaic.h"
#include <bits/stdc++.h>
using namespace std;
 
std::vector<long long> mosaic(std::vector<int> X, std::vector<int> Y,
                              std::vector<int> T, std::vector<int> B,
                              std::vector<int> L, std::vector<int> R)
{
    int N = X.size();
    int Q = T.size();
    if(N == 0) {
        return vector<long long>(Q, 0LL);
    }
 
    // Step 1: fill A (N*N array, store as char 0/1)
    vector<unsigned char> A;
    A.resize((size_t)N * (size_t)N);
    // fill top row
    for(int j=0; j<N; j++) {
        A[0*N + j] = X[j];
    }
    // fill left column
    for(int i=0; i<N; i++) {
        A[i*N + 0] = Y[i];
    }
    // fill rest
    for(int i=1; i<N; i++) {
        for(int j=1; j<N; j++) {
            A[i*N + j] = (1 - A[(i-1)*N + j]) * (1 - A[i*N + j-1]);
        }
    }
 
    // Step 2: compute prefix sums in P (N+1)*(N+1) array
    vector<int> P;
    P.resize((size_t)(N+1) * (size_t)(N+1));
    // initialize P with 0's
    std::fill(P.begin(), P.end(), 0);
 
    for(int i=0; i<N; i++) {
        int baseRow = (i+1)*(N+1);
        int prevRow = i*(N+1);
        for(int j=0; j<N; j++) {
            P[baseRow + (j+1)] = P[prevRow + (j+1)] + P[baseRow + j] - P[prevRow + j] + (A[i*N + j]);
        }
    }
 
    // Step 3: answer queries using P
    vector<long long> C;
    C.resize(Q);
    for(int k=0; k<Q; k++) {
        int top = T[k];
        int bottom = B[k];
        int left = L[k];
        int right = R[k];
        long long blackCount = (long long)P[(bottom+1)*(N+1) + (right+1)]
            - (long long)P[(top)*(N+1) + (right+1)]
            - (long long)P[(bottom+1)*(N+1) + left]
            + (long long)P[(top)*(N+1) + left];
        C[k] = blackCount;
    }
 
    return C;
}
\end{lstlisting}

\subsubsection{Submission 2}
The model scored 20 points in the second submission on subtasks 1, 3, and 5.
\begin{lstlisting}[language=C++]
#include "mosaic.h"
#include <bits/stdc++.h>
using namespace std;
 
// We'll implement the mosaic function as specified.
std::vector<long long> mosaic(std::vector<int> X, std::vector<int> Y,
                              std::vector<int> T, std::vector<int> B,
                              std::vector<int> L, std::vector<int> R)
{
    int N = X.size();
    int Q = T.size();
 
    // If N=1, the mosaic is just tile(0,0), queries are easy.
    if (N == 1) {
        vector<long long> result(Q);
        for (int q = 0; q < Q; q++) {
            // subrectangle T..B x L..R is always T=0,B=0,L=0,R=0 if we have queries.
            if (T[q] == 0 && B[q] == 0 && L[q] == 0 && R[q] == 0) {
                // black tile count is 1 if X[0]=1 or Y[0]=1, and X[0]=Y[0], else 0.
                // The problem states X[0] = Y[0], so tile(0,0) is black if X[0]=1.
                if (X[0] == 1) result[q] = 1; else result[q] = 0;
            } else {
                // no other queries are possible if N=1, but if they are, the subrectangle is empty or out of range.
                result[q] = 0;
            }
        }
        return result;
    }
 
    // Precompute prefix sums for X and Y.
    vector<long long> prefixX(N+1, 0), prefixY(N+1, 0);
    for (int i = 1; i <= N; i++) {
        prefixX[i] = prefixX[i-1] + X[i-1];
        prefixY[i] = prefixY[i-1] + Y[i-1];
    }
 
    // Determine alpha mod 2 and beta mod 2 from tile(1,1).
    // C[1,1] = 1 if X[1] and Y[1] are 0, else 0.
    int c11 = 0;
    if (N > 1) {
        if (X.size() > 1 && Y.size() > 1 && X[1] == 0 && Y[1] == 0) c11 = 1; else c11 = 0;
    } else {
        c11 = 0; // but N=1 scenario is handled above.
    }
 
    int alphaMod2, betaMod2;
    if (c11 == 1) {
        alphaMod2 = 0;
        betaMod2 = 0;
    } else {
        alphaMod2 = 0;
        betaMod2 = 1;
    }
 
    // define functions to count even and odd in a range.
    auto countEvenInRange = [&](int a, int b){
        if (b < a) return 0LL;
        long long A = a; long long B = b;
        long long evenCount = ((long long)B + 2) / 2 - ((long long)A + 1) / 2;
        return evenCount;
    };
 
    auto countOddInRange = [&](int a, int b){
        if (b < a) return 0LL;
        long long A = a; long long B = b;
        long long oddCount = ((long long)B + 1) / 2 - (long long)A / 2;
        return oddCount;
    };
 
    // Now for each query, compute the black tile count.
    std::vector<long long> result(Q);
    for (int q = 0; q < Q; q++) {
        long long t = T[q];
        long long b = B[q];
        long long l = L[q];
        long long r = R[q];
 
        long long count = 0;
 
        // if T=0, add black tile from row0 and columns L..R.
        if (t == 0 && r >= l) {
            long long sumRow0 = prefixX[r+1] - prefixX[(int)l]; // sum of X[L..R]
            count += sumRow0;
        }
 
        // if L=0, add black tile from column0 and rows T..B.
        if (l == 0 && b >= t) {
            long long sumCol0 = prefixY[b+1] - prefixY[(int)t]; // sum of Y[T..B]
            count += sumCol0;
        }
 
        // if T=0 and L=0, subtract tile(0,0) if black.
        if (t == 0 && l == 0 && r >= 0 && b >= 0) {
            if (X[0] == 1) {
                count -= 1;
            }
        }
 
        // add black tile from the interior submatrix T'=max(T,1)..B x L'=max(L,1)..R.
        int Tprime = max((int)t, 1);
        int Lprime = max((int)l, 1);
        if (Tprime <= b && Lprime <= r) {
            long long evenI = countEvenInRange(Tprime, (int)b);
            long long oddI = countOddInRange(Tprime, (int)b);
            long long evenJ = countEvenInRange(Lprime, (int)r);
            long long oddJ = countOddInRange(Lprime, (int)r);
 
            long long interiorCount = 0;
            if (alphaMod2 == betaMod2) {
                // black tile appear when i%
                interiorCount = evenI * evenJ + oddI * oddJ;
            } else {
                // black tile appear when i%
                interiorCount = evenI * oddJ + oddI * evenJ;
            }
            count += interiorCount;
        }
 
        result[q] = count;
    }
 
    return result;
}
\end{lstlisting}

%% file: samples/sphinx.tex
\subsection{Sphinx}
A total of 71.5 points were scored on Sphinx across two separate submissions.

\subsubsection{Submission 1}
The first submission achieved a score of 50 out of 100, with 50\% partial credit earned on all subtasks.

\begin{lstlisting}[language=C++]
#include "sphinx.h"
#include <bits/stdc++.h>
int perform_experiment(std::vector<int> E);
std::vector<int> find_colours(int Nx, std::vector<int> Xx, std::vector<int> Yx) {
    int N = Nx;
    std::vector<std::vector<int>> adj(N);
    int M = Xx.size();
    for (int i = 0; i < M; i++) {
        int u = Xx[i], v = Yx[i];
        adj[u].push_back(v);
        adj[v].push_back(u);
    }
 
    struct UnionFind {
        std::vector<int> parent, rank;
        UnionFind(int n) : parent(n), rank(n, 0) {
            for (int i = 0; i < n; i++) parent[i] = i;
        }
        int findp(int x) {
            if (parent[x] == x) return x;
            parent[x] = findp(parent[x]);
            return parent[x];
        }
        void unite(int x, int y) {
            x = findp(x); y = findp(y);
            if (x == y) return;
            if (rank[x] < rank[y]) parent[x] = y;
            else if (rank[x] > rank[y]) parent[y] = x;
            else {parent[y] = x; rank[x]++;}
        }
    } uf(N);
 
    auto connectedComponentsInSubset = [&](const std::vector<int>& subset) {
        static std::vector<int> visited;
        visited.assign(N, -1);
        for (int v : subset) visited[v] = 0;
        int count = 0;
        for (int v : subset) if (visited[v] == 0) {
            count++;
            std::queue<int> q;
            q.push(v);
            visited[v] = 1;
            while(!q.empty()) {
                int u = q.front(); q.pop();
                for (int w : adj[u]) {
                    if (visited[w] == 0) {
                        visited[w] = 1;
                        q.push(w);
                    }
                }
            }
        }
        return count;
    };
 
    auto queryMonochromComps = [&](const std::vector<int>& S) {
        std::vector<int> Earr(N);
        static std::vector<int> visited2;
        visited2.assign(N, -1);
        std::vector<int> Sset(N, 0);
        for (int v : S) Sset[v] = 1;
        for (int i = 0; i < N; i++) Earr[i] = (Sset[i] == 1) ? -1 : N;
 
        int res = perform_experiment(Earr);
 
        std::vector<int> compInput; compInput.reserve(N - S.size());
        for (int i = 0; i < N; i++) if (Sset[i] == 0) compInput.push_back(i);
        int compCount = 0;
        if (!compInput.empty()) compCount = connectedComponentsInSubset(compInput);
        else compCount = 0;
 
        int monos = res - compCount;
        return monos;
    };
 
    std::vector<int> c(N+1, 0);
    c[0] = 0; // no vertices, #connectedComponents(0) = 0.
    for (int k = 1; k <= N; k++) {
        std::vector<int> S; S.reserve(k);
        for (int i = 0; i < k; i++) S.push_back(i);
        c[k] = queryMonochromComps(S);
    }
 
    for (int k = 1; k <= N; k++) {
        int v = k-1;
        int merges = c[k-1] + 1 - c[k];
        if (merges < 0) merges = 0; // safety.
 
        auto findMerges = [&](auto &self, std::vector<int>& roots, int merges) -> void {
            if (merges == 0) return;
            if (merges == (int)roots.size()) {
                for (int r : roots) {
                    uf.unite(v, r);
                }
                return;
            } else {
                if (roots.size() == merges) {
                    for (int r : roots) {
                        uf.unite(v, r);
                    }
                    return;
                } else {
                    int half = roots.size()/2;
                    std::vector<int> left(roots.begin(), roots.begin() + half);
                    std::vector<int> right(roots.begin() + half, roots.end());
 
                    std::vector<int> S2; S2.reserve(1 + v);
                    S2.push_back(v);
                    for (int r : left) {
                        for (int i = 0; i < v; i++) {
                            if (uf.findp(i) == r) {
                                S2.push_back(i);
                            }
                        }
                    }
 
                    std::vector<int> Earr(N);
                    std::vector<int> Sset2(N, 0);
                    for (int i : S2) Sset2[i] = 1;
                    for (int i = 0; i < N; i++) Earr[i] = (Sset2[i] == 1) ? -1 : N;
 
                    int res = perform_experiment(Earr);
 
                    std::vector<int> compInput; compInput.reserve(N - S2.size());
                    static std::vector<int> visited2; visited2.assign(N, -1);
                    for (int i = 0; i < N; i++) visited2[i] = 0;
                    for (int i : S2) visited2[i] = -1;
                    compInput.clear();
                    for (int i = 0; i < N; i++) if (visited2[i] == 0) compInput.push_back(i);
                    int compCount = 0;
                    if (!compInput.empty()) {
                        for (int ip : compInput) visited2[ip] = 0; // re-mark for BFS.
                        for (int ip : compInput) if (visited2[ip] == 0) {
                            compCount++;
                            std::queue<int> q;
                            q.push(ip);
                            visited2[ip] = 1;
                            while(!q.empty()) {
                                int u = q.front(); q.pop();
                                for (int w : adj[u]) {
                                    if (visited2[w] == 0) {
                                        visited2[w] = 1;
                                        q.push(w);
                                    }
                                }
                            }
                        }
                    } else compCount = 0;
 
                    int monos2 = res - compCount;
                    int mergesInLeft = left.size() + 1 - monos2;
 
                    self(self, left, mergesInLeft);
                    self(self, right, merges - mergesInLeft);
                }
            }
        };
 
        std::set<int> distinctRoots;
        for (int i = 0; i < v; i++) distinctRoots.insert(uf.findp(i));
        std::vector<int> rootsVec(distinctRoots.begin(), distinctRoots.end());
 
        findMerges(findMerges, rootsVec, merges);
    }
 
    std::vector<int> rootID(N, -1);
    std::vector<int> colorID(N, -1);
    int nextID = 0;
    for (int i = 0; i < N; i++) {
        int r = uf.findp(i);
        if (rootID[r] == -1) {
            rootID[r] = nextID++;
        }
        colorID[i] = rootID[r];
    }
 
    return colorID;
}
\end{lstlisting}

\subsubsection{Submission 2}
Submission 2 achieved 43 points on subtasks 1, 2, and 3.
\begin{lstlisting}[language=C++]
#include "sphinx.h"
#include <bits/stdc++.h>
using namespace std;
 
int perform_experiment(vector<int> E);
 
vector<int> find_colours(int N, vector<int> X, vector<int> Y) {
    // We'll store adjacency.
    vector<vector<int>> adj(N);
    vector<vector<bool>> adjMatrix(N, vector<bool>(N, false));
    for (int i = 0; i < (int)X.size(); i++) {
        int u = X[i];
        int v = Y[i];
        adj[u].push_back(v);
        adj[v].push_back(u);
        adjMatrix[u][v] = adjMatrix[v][u] = true;
    }
 
    vector<int> final_colors(N, -1);
    vector<int> unassigned; for (int i = 0; i < N; i++) unassigned.push_back(i);
 
    // define group testing functions.
    // function query_count_in_T: how many in T have color x.
    auto query_count_in_T = [&](const vector<int> &T, int x) {
        vector<int> E(N, x);
        vector<int> inT(N, 0);
        for (int v : T) {
            E[v] = -1;
            inT[v] = 1;
        }
 
        int R = perform_experiment(E);
 
        // number c of connected components in the rest.
        vector<int> complement;
        for (int i = 0; i < N; i++) if (!inT[i]) complement.push_back(i);
        // BFS for c.
        vector<int> visited(N, 0);
        int ccount = 0;
        for (int v : complement) {
            if (!visited[v]) {
                ccount++;
                stack<int> st;
                st.push(v);
                visited[v] = 1;
                while (!st.empty()) {
                    int node = st.top(); st.pop();
                    for (int nb : adj[node]) {
                        if (!inT[nb] && !visited[nb]) {
                            visited[nb] = 1;
                            st.push(nb);
                        }
                    }
                }
            }
        }
 
        int c = ccount;
 
        int countx = (int)T.size() - (R - c);
        return countx;
    };
 
    // function find vertices in T with color x.
    auto find_vertices_in_T_with_color_x = [&](vector<int> T, int x) {
        vector<int> result;
        vector<int> S = T;
        int count = query_count_in_T(S, x);
        while (count > 0) {
            int l = 0; int r = S.size() - 1;
            while (l < r) {
                int mid = (l + r) / 2;
                vector<int> part(S.begin() + l, S.begin() + mid + 1);
                int cpart = query_count_in_T(part, x);
                if (cpart > 0) {
                    r = mid;
                } else {
                    l = mid + 1;
                }
            }
            int found = S[l];
            result.push_back(found);
 
            // remove found from S.
            auto it = find(S.begin(), S.end(), found);
            if (it != S.end()) S.erase(it);
 
            count = query_count_in_T(S, x);
        }
        return result;
    };
 
    auto find_independent_set_heuristic = [&](vector<int> &verts) {
        vector<int> best;
        // We'll use a random approach.
        for (int rep = 0; rep < 10; rep++) {
            vector<int> attempt;
            vector<int> local = verts;
            random_shuffle(local.begin(), local.end());
            for (int v : local) {
                bool can = true;
                for (int u : attempt) {
                    if (adjMatrix[u][v]) {
                        can = false;
                        break;
                    }
                }
                if (can) attempt.push_back(v);
            }
            if (attempt.size() > best.size()) best = attempt;
        }
        return best;
    };
 
    vector<int> U = unassigned;
    while (!U.empty()) {
        vector<int> T = find_independent_set_heuristic(U);
 
        vector<int> assignedInT;
        vector<int> S = T;
        while (!S.empty()) {
            bool foundX = false;
            for (int x = 0; x < N; x++) {
                if (S.empty()) break;
                int countx = query_count_in_T(S, x);
                if (countx > 0) {
                    vector<int> found = find_vertices_in_T_with_color_x(S, x);
                    for (int v : found) {
                        final_colors[v] = x;
                        assignedInT.push_back(v);
                    }
                    for (int v : found) {
                        auto it = find(S.begin(), S.end(), v);
                        if (it != S.end()) S.erase(it);
                    }
                    foundX = true;
                }
            }
            if (!foundX) {
                // If no color found for S, we can't assign those vertices.
                // But maybe it's impossible.
                break;
            }
        }
 
        // remove T from U.
        set<int> assignedSet(assignedInT.begin(), assignedInT.end());
        vector<int> newU;
        for (int v : U) {
            if (assignedSet.find(v) == assignedSet.end()) newU.push_back(v);
        }
        U = newU;
    }
 
    return final_colors;
}
\end{lstlisting}

%% file: main.bbl
\begin{thebibliography}{10}

\bibitem{austin2021program}
Jacob Austin, Augustus Odena, Maxwell Nye, Maarten Bosma, Henryk Michalewski,
  David Dohan, Ellen Jiang, Carrie Cai, Michael Terry, Quoc Le, et~al.
\newblock Program synthesis with large language models.
\newblock {\em arXiv preprint arXiv:2108.07732}, 2021.

\bibitem{chen2021evaluating}
Mark Chen, Jerry Tworek, Heewoo Jun, Qiming Yuan, Henrique Ponde De~Oliveira
  Pinto, Jared Kaplan, Harri Edwards, Yuri Burda, Nicholas Joseph, Greg
  Brockman, et~al.
\newblock Evaluating large language models trained on code.
\newblock {\em arXiv preprint arXiv:2107.03374}, 2021.

\bibitem{deepseekai2025deepseekr1incentivizingreasoningcapability}
DeepSeek-AI, Daya Guo, Dejian Yang, Haowei Zhang, Junxiao Song, Ruoyu Zhang,
  et~al.
\newblock Deepseek-r1: Incentivizing reasoning capability in llms via
  reinforcement learning.
\newblock {\em arXiv preprint arXiv:2501.12948}, 2025.

\bibitem{jaech2024openai}
Aaron Jaech, Adam Kalai, Adam Lerer, Adam Richardson, Ahmed El-Kishky, Aiden
  Low, Alec Helyar, Aleksander Madry, Alex Beutel, Alex Carney, et~al.
\newblock Openai o1 system card.
\newblock {\em arXiv preprint arXiv:2412.16720}, 2024.

\bibitem{jimenez2023swe}
Carlos~E Jimenez, John Yang, Alexander Wettig, Shunyu Yao, Kexin Pei, Ofir
  Press, and Karthik Narasimhan.
\newblock Swe-bench: Can language models resolve real-world github issues?
\newblock {\em arXiv preprint arXiv:2310.06770}, 2023.

\bibitem{leblond2023alphacode}
{Leblond, R{\'e}mi and Gimeno, Felix and Altch{\'e}, Florent and Saade, Alaa
  and Ruddock, Anton and Tallec, Corentin and Powell, George and Grill,
  Jean-Bastien and Miku{\l}a, Maciej and Lochbrunner, Matthias and others}.
\newblock Alphacode 2 technical report.
\newblock
  \url{https://storage.googleapis.com/deepmind-media/AlphaCode2/AlphaCode2_Tech_Report.pdf},
  December 2023.
\newblock Accessed: 2025-01-14.

\bibitem{li2022competition}
Yujia Li, David Choi, Junyoung Chung, Nate Kushman, Julian Schrittwieser,
  R{\'e}mi Leblond, Tom Eccles, James Keeling, Felix Gimeno, Agustin Dal~Lago,
  et~al.
\newblock Competition-level code generation with alphacode.
\newblock {\em Science}, 378(6624):1092--1097, 2022.

\bibitem{cf_rating_1}
Mike Mirzayanov.
\newblock Codeforces rating system.
\newblock \url{https://codeforces.com/blog/entry/102}, 2010.

\bibitem{cf_rating_2}
Mike Mirzayanov.
\newblock Open codeforces rating system.
\newblock \url{https://codeforces.com/blog/entry/20762}, 2016.

\bibitem{cf_rating_3}
Mike Mirzayanov.
\newblock Codeforces: Soon we will change the rating calculation for new
  accounts.
\newblock \url{https://codeforces.com/blog/entry/77890}, 2020.

\bibitem{openai_swe_bench_verified}
{OpenAI}.
\newblock Introducing swe-bench verified.
\newblock \url{https://openai.com/index/introducing-swe-bench-verified/},
  August 2024.
\newblock Accessed: 2025-01-14.

\bibitem{openai_learning_reason_llms}
{OpenAI}.
\newblock Learning to reason with llms.
\newblock \url{https://openai.com/index/learning-to-reason-with-llms/},
  September 2024.
\newblock Accessed: 2025-01-14.

\bibitem{openaio3}
OpenAI.
\newblock Openai o3 system card.
\newblock {\em Technical Report}, 2025.

\bibitem{schick2023toolformer}
Timo Schick, Jane Dwivedi-Yu, Roberto Dess{\`\i}, Roberta Raileanu, Maria
  Lomeli, Eric Hambro, Luke Zettlemoyer, Nicola Cancedda, and Thomas Scialom.
\newblock Toolformer: Language models can teach themselves to use tools.
\newblock {\em Advances in Neural Information Processing Systems},
  36:68539--68551, 2023.

\bibitem{kimiteam2025kimik15scalingreinforcement}
Kimi Team, Angang Du, Bofei Gao, Bowei Xing, Changjiu Jiang, Cheng Chen, Cheng
  Li, Chenjun Xiao, et~al.
\newblock Kimi k1.5: Scaling reinforcement learning with llms.
\newblock {\em arXiv preprint arXiv:2501.12599}, 2025.

\bibitem{wei2022chain}
Jason Wei, Xuezhi Wang, Dale Schuurmans, Maarten Bosma, Fei Xia, Ed~Chi, Quoc~V
  Le, Denny Zhou, et~al.
\newblock Chain-of-thought prompting elicits reasoning in large language
  models.
\newblock {\em Advances in neural information processing systems},
  35:24824--24837, 2022.

\end{thebibliography}
